\documentclass[a4paper,12pt]{article}
\usepackage{geometry}
\usepackage{type1cm}        % activate if the above 3 fonts are
\usepackage{graphicx}        % standard LaTeX graphics tool
\usepackage[bottom]{footmisc}% places footnotes at page bottom

\usepackage{amssymb}
\usepackage{amsmath}
\usepackage{bm}
\usepackage{mathtools}
\usepackage{xcolor}
\usepackage{tikz}
\usepackage{thmtools}

\usepackage{nicefrac}

\usepackage{amsthm}
\theoremstyle{definition}
\newtheorem{examplex}{Example}
\newenvironment{example}
  {\pushQED{\qed}\examplex}
  {\popQED\endexamplex}
\newtheorem{definition}{Definition}
\newtheorem{theorem}{Theorem}
\newtheorem{lemma}{Lemma}
\newtheorem{proposition}{Proposition}
\newtheorem{corollary}{Corollary}

\usepackage[shortlabels]{enumitem}

\usepackage[hidelinks]{hyperref}

\DeclarePairedDelimiter{\group}{(}{)}

\DeclarePairedDelimiter{\set}{\{}{\}}

\DeclarePairedDelimiter{\abs}{\vert}{\vert}

\newcommand{\things}{\mathcal{T}}
\newcommand{\setsofthings}{\mathcal{P}(\things)}
\newcommand{\finsetsofthings}{\mathcal{P}_\mathrm{fin}(\things)}

\newcommand{\options}{\mathcal{O}}

\newcommand{\states}{\mathcal{X}}
\newcommand{\rewards}{\mathcal{R}}
\newcommand{\statesandrewards}{\states\times\rewards}

\newcommand{\sodos}[1][]{K_{#1}}
\newcommand{\sodoss}{\mathbf{K}}

\newcommand{\naturals}{\mathbb{N}}

\newcommand{\reals}{\mathbb{R}}

\newcommand{\opt}[1][]{u_{#1}}

\newcommand{\opts}{\mathcal{V}}

\newcommand{\optset}[1][]{A_{#1}}

\newcommand{\assessment}{\mathcal{A}}

\newcommand{\desirset}[1][]{D_{#1}}

\newcommand{\desirsets}{\mathbf{D}}

\newcommand{\rejectset}[1][]{K_{#1}}

\newcommand{\setofdesirsets}{\mathcal{D}}

\newcommand{\ddualopts}[1][]{\opts^\circ}

\newcommand{\ldualopts}[1][]{\underline{\opts}^\ast}

\newcommand{\nml}[1][{\opt[o]}]{\normalise_{#1}}
\newcommand{\cset}[3][]{\set[#1]{#2\colon#3}}

\DeclareMathOperator{\posi}{posi}

\DeclareMathOperator{\chull}{CH}

\DeclareMathOperator{\closure}{cl}
\DeclareMathOperator{\fin}{fin}
\DeclareMathOperator{\trans}{trans}

\DeclareMathOperator{\iden}{id}

\begin{document}

%\title*{An axiomatic characterisation for choice functions that are based on sets of strict partial orders}
%\title*{Proper choice functions}
%\subtitle{An axiomatic foundation for decision making based on sets of strict partial orders}
% \title{A theory of desirable things,\\ and its application to choice functions}
\title{A theory of desirable things}
%\title*{Coherent choice functions that allow for infinite option sets}b 
% Use \titlerunning{Short Title} for an abbreviated version of
% your contribution title if the original one is too long
\author{Jasper De Bock}
\date{}
% Use \authorrunning{Short Title} for an abbreviated version of
% your contribution title if the original one is too long

%
% Use the package "url.sty" to avoid
% problems with special characters
% used in your e-mail or web address
%
\maketitle

% \abstract*{Each chapter should be preceded by an abstract (no more than 200 words) that summarizes the content. The abstract will appear \textit{online} at \url{www.SpringerLink.com} and be available with unrestricted access. This allows unregistered users to read the abstract as a teaser for the complete chapter.
% Please use the 'starred' version of the \texttt{abstract} command for typesetting the text of the online abstracts (cf. source file of this chapter template \texttt{abstract}) and include them with the source files of your manuscript. Use the plain \texttt{abstract} command if the abstract is also to appear in the printed version of the book.}

\begin{abstract}
Inspired by the theory of desirable gambles that is used to model uncertainty in the field of imprecise probabilities, I present a theory of desirable things. Its aim is to model a subject's beliefs about which things are desirable. What the things are is not important, nor is what it means for them to be desirable. It can be applied to gambles, calling them desirable if a subject accepts them, but it can just as well be applied to pizzas, calling them desirable if my friend Arthur likes to eat them. Other useful examples of things one might apply this theory to are propositions, horse lotteries, or preferences between any of the above. 
Regardless of the particular things that are considered, inference rules are imposed by means of an abstract closure operator, and models that adhere to these rules are called coherent. I consider two types of models, each of which can capture a subject's beliefs about which things are desirable: sets of desirable things and sets of desirable sets of things. A crucial result is that the latter type can be represented by a set of the former.
%To illustrate this result, I show how it can be used to establish a correspondence between choice functions and sets of preference orders. In this way, I recover and extend existing axiomatic characterisations of various imprecise-probabilistic decision making methods.
\end{abstract}

\section{Introduction}

The theory of imprecise probabilities~\cite{augustin2013:itip,walley1991} is often thought of as a theory of partially specified probabilities, which involves manipulating sets of probabilities and their lower and upper expectations. Its mathematical underpinnings, however, are provided by an underlying theory of sets of desirable gambles~\cite{walley1991,williams1975,walley2000,couso2011:desirable,quaeghebeur2012:itip}: sets of gambles---rewards with an uncertain payoff---that a subject finds desirable, in the sense that she prefers those gambles to the status quo---to the trivial gamble with zero payoff. Rewards are typically taken to be expressed in units of some linear utility scale, and this them implies that positive linear combinations of desirable gambles are desirable themselves. Sets of desirable gambles that satisfy this condition (as well as some other, less essential conditions) are called coherent. Due to the geometric nature of the coherence conditions, inference with desirable gambles is typically simple and intuitive, a feature that is particularly handy, also when it comes to designing proofs. Most crucially, however, well known imprecise probability models such as credal sets (closed convex sets of probabilites), lower and upper expectations (or previsions), partial preference oderings, belief functions and lower and upper probabilities, all correspond to special cases of coherent sets of desirable gambles~\cite{walley2000}, which explains the importance of the latter as a basis for imprecise-probabilistic reasoning.

When it comes to decision making, their connection with partial preference orderings makes sets of desirable gambles an important tool for decision making as well. Decision making by maximising expected utility, for example, can be seen as a special case of decision making with sets of desirable gambles. Nevertheless, decision making with sets of desirable gambles has its limitations. Most importantly, it is binary in nature, meaning that all decisions can be reduced to pairwise choices between gambles~\cite{seidenfeld2010}. To address this issue, the theory of sets of desirable gambles has been extended to sets of desirable sets of gambles~\cite{debock2018,pmlr-v103-de-bock19b,VANCAMP2023390}, where a set of gambles is said to be desirable if it contains at least one desirable gamble. So essentially, this generalisation adds a notion of disjunction to the theory, allowing statements such as ``one of these two gambles is desirable''. Interestingly, it turns out that this alllows the theory to capture non-binary decision making, through an equivalence with so-called coherent choice functions. A crucial result in that respect, is that every coherent set of desirable sets of gambles can be equivalently represented by a set of coherent sets of desirable gambles~\cite{debock2018}. That is, this generalised theory of desirable gambles simply amounts to using sets of the more simple models in the original theory. Through the connection between sets of desirable sets of gambles and choice functions on the one hand, and sets of desirable gambles and other imprecise probability models on the other hand, this leads to axiomatic characterisations for various types of decision making, such as, for example, decision making using (non necessarily convex) sets of probabilities~\cite{pmlr-v103-de-bock19b,ipmu2020debock:arxiv}

The main contribution of this paper is to show that many of the central ideas behind the theory of desirable gambles, both in its original and generalised form, are not constrained to the context of gambles. We do this by putting forward a theory of desirable things. As we will see, these things can be, quite literally, anything, even pizzas. 
All that is needed to develop the theory is an inference mechanism that allows us to infer the desirability of things from that of other things; we do this by means of an abstract closure operator. In the case of desirable gambles, this would typically be closure with respect to positive linear combinations. The point of the paper, however, is that any other closure operator can be used instead, and that the gambles can instead by arbitrary things. This allows us to generalise the notion of a coherent set of desirable gambles to that of a coherent set of desirable things, and similarly for coherent sets of desirable sets of gambles and coherent sets of desirable sets of things. Furthermore, and perhaps most surprisingly, we show that the connection between these two types of models also extends from gambles to things.

We start in Section~\ref{sec:desirablethings} by introducing the two types of models that we consider: sets of desirable things and sets of desirable sets of things. The rules that these models should adhere to are discussed in Section~\ref{sec:coherence}, including how inference can be expressed by means of an abstract closure operator. This leads to a notion of coherence for both types of models. Next, in Section~\ref{sec:representation}, we explain how these two types of models are connected: sets of desirable things can be seen as special cases of sets of desirable sets of things, and most importantly, the latter can furthermore always be represented by a set of the latter. The last main section of the paper, Section~\ref{sec:finitecase}, investigates to which extent the coherence axioms can be further simplified, without giving up on the representation result that connects both types of models. We establish several conditions under which this is possible, which typically involve the use of closure operators that satisfy additional properties. We also see that some of these simplifications require that we focus solely on (representing) those desirable sets of things that are finite.

\section{Desirable things}\label{sec:desirablethings}

Let $\mathcal{T}$ be a set containing all things whose desirability we wish to model. We'll use $\mathcal{P}(\mathcal{T})$ to denote its powerset: the set of all sets of things. A crucial feature of our framework, is that it does not matter what the things in $\things$ are, nor what it means for them to be desirable. Desirability is simply an abstract feature that these things may or may not have, and the goal is to model a subject's beliefs about which of the things in $\things$ are desirable. In particular, we put forward two different types of models that can be used for this purpose.

The first---and most simple---model is a \emph{set of desirable things} $D\subseteq\mathcal{T}$, which we will sometimes refer to as an SDT. As the terminology suggests, this is simply a set containing things that our subject deems desirable. No exhaustivity claim is made though: the model does not claim that the things in $D$ are the only desirable ones; it simply says that every thing $t$ in $D$ is deemed desirable.

\begin{example}
Let $\mathcal{T}$ be the set of all types of pizza, and call a pizza desirable if my friend Arthur likes to eat them. Let the subject who's beliefs we are modelling be myself. A set of desirable things---in this case, a set of desirable pizzas---is then simply a set $D\subseteq\mathcal{T}$ of pizzas that I think Arthur likes to eat. I might for example provide the set of desirable pizzas
\begin{equation*}
D=\{\mathrm{peperoni},\mathrm{meatballs}\},
\end{equation*}
which expresses that I think that Arthur likes to eat pizza peperoni and pizza meatballs.
\end{example}

The second model is a \emph{set of desirable sets of things} $K\subseteq\mathcal{P}(\mathcal{T})$. Rather than use desirable things as its elements, this model instead considers \emph{desirable sets of things}, which are \emph{sets that contain at least one desirable thing}. For ease of reference, we will often simply speak of \emph{desirable sets}, leaving it implicit that they are actually desirable sets of things, and then refer to $K$ as a set of desirable sets\footnote{If the things we are considering are sets themselves, this is a bad idea, as the distinction between sets of desirable things and sets of desirabe sets then becomes unclear. In such a case, it seems preferable to not use the shorthand sets of desirable sets, and always speak of sets of desirable sets of things.} or, even shorter, as an SDS. Again, no exhaustivity claim is made: there might be desirable sets that are not in $K$. The model simply---and only---says that every set $T$ in $K$ is deemed desirable by the subject who's beliefs we are modelling, meaning that she thinks that every set $T\in K$ contains at least one desirable thing. This second type of model is more complex, but also has more expressive power. The following two examples provide a first illustration.

\begin{example}
Continuing with the pizza example, consider the following problem: I remember that I've been to a restaurant with Arthur, and that he ordered a pizza there that he liked a lot. I don't remember which pizza it was though. On the menu of the restaurant, there are two pizzas: pizza peperoni and olives, and pizza chicken and olives. So I know that Arthur likes at least one of these two pizzas or, using our abstract terminology, that at least one of these two pizzas is desirable. I would like to include this piece of information in my model. With sets of desirable things, this is not possible. Sets of desirable sets, however, can. It suffices to assess that the set
\begin{equation*}
T=\{\mathrm{peperoni~and~olives}, \mathrm{chicken~and~olives}\}
\end{equation*}
is desirable, meaning that it contains at least one desirable pizza. I could for example put forward the set of desirable sets
\begin{equation*}
K=\big\{\{\mathrm{peperoni}\},\{\mathrm{meatballs}\},\{\mathrm{peperoni~and~olives}, \mathrm{chicken~and~olives}\}\big\},
\end{equation*}
which contains three desirable sets. The first one is $\{\mathrm{peperoni}\}$. Since at least one of the pizzas is desirable, this simply expresses that I think Arthur likes pizza peperoni. Similarly, the second set, $\{\mathrm{meatballs}\}$, expresses that I think Arthur likes pizza meatballs. The third desirable set in $K$ is the set $T$ that we considered before. It expresses that I think Arthur likes pizza peperoni and olives \emph{or} pizza chicken and olives, but that I don't know which of these two he likes.
\end{example}

\begin{example}
Let $\mathcal{T}$ be some set of propositions and let desirable propositions be propositions that are true. Consider now three propositions $p_1$, $p_2$ and $p_3$ in $\mathcal{T}$ . I furthermore know that at least two of these propositions are true, but I don't know which two are true. I can model this information with the following set of desirable sets (of propositions):
\begin{equation*}
K=\big\{\{p_1,p_2\},\{p_2,p_3\},\{p_1,p_3\}\big\}.
\end{equation*}
It expresses that if I remove any of the three considered propositions, then---since at least two of the propositions are true---the set of the remaining two propositions will contain at least one proposition that is true.
\end{example}

\section{Coherence}\label{sec:coherence}

For many sets of things $\mathcal{T}$, and many notions of desirability, it makes sense to impose rules that statements about desirability should adhere to. One might think of them as rationality constraints. We wil consider three such rules here. The first rule puts forward a set $A_{\mathrm{not}}\subseteq\mathcal{T}$ of things that should never---on grounds of rationality---be desirable:
\begin{enumerate}[label=$\mathrm{R_{not}}$.,ref=$\mathrm{R_{not}}$,leftmargin=*]
\item The things in $A_{\mathrm{not}}$ are not desirable.\label{rulenot}
\end{enumerate}
The second puts forward a set $A_\mathrm{des}\subseteq\things$ of things that should always be desirable:
\begin{enumerate}[label=$\mathrm{R_{des}}$.,ref=$\mathrm{R_{des}}$,leftmargin=*]
\item The things in $A_\mathrm{des}$ are desirable.\label{ruledes}
\end{enumerate}
% This second rule includes as a special case the rule that every thing in some set $A_\mathrm{des}\subseteq\things$ should be desirable. This corresponds to letting $\mathcal{A}_\mathrm{des}\coloneqq\{\{t\}\colon t\in A_\mathrm{des}\}$. In such cases, we will simply specify $A_\mathrm{des}$ instead of $\mathcal{A}_\mathrm{des}$.

\begin{example}
Continuing with the pizza example, we might want to impose, on grounds of rationality, that pizza Hawaii should never be desirable and that pizza margherita should always be desirable. To do so, it suffices to let $A_\mathrm{not}=\{\mathrm{Hawaii}\}$ and $A_\mathrm{des}=\{\mathrm{margherita}\}$
\end{example}

Our third rule builds in an inference mechanism, expressed by means of a closure operator $\closure$.

\begin{definition}[Closure operator]
\label{def:closop}
A map $\closure\colon\mathcal{P}(\mathcal{T})\to\mathcal{P}(\mathcal{T})$ is a closure operator\footnote{\label{footnote:logic}Similar operators are also used in abstract logic. An operator that satisfies \ref{ax:clos:ext}--\ref{ax:clos:idem} is then called a consequence operator~\cite{wojcicki1988}, and a consequence operator that additionally satisfies \ref{ax:clos:empty} is called axiomless \cite{los1958}.} if
\begin{enumerate}[label=$\mathrm{cl}_{\arabic*}$.,ref=$\mathrm{cl}_{\arabic*}$,leftmargin=*]
\item\label{ax:clos:ext} $A\subseteq\closure(A)$\hfill [extensive]
\item\label{ax:clos:mon} if $A\subseteq B$, then $\closure(A)\subseteq\closure(B)$\hfill [monotone]
\item\label{ax:clos:idem} $\closure(\closure(A))=\closure(A)$\hfill [idempotent]
\item\label{ax:clos:empty}
$\closure(\emptyset)=\emptyset$
\end{enumerate}
\end{definition}
These conditions are slightly stronger than what is usually required of a closure operator. In particular, the last condition, stating that the closure of the empty set should be empty, is typically not imposed. We choose to impose it anyway because it better corresponds to the interpretation that we want to attach to such a closure operator, which goes as follows: for every set of things $A$, we take $\closure(A)$ to contain all the things whose desirability can be inferred from the desirability of the things in $A$:

\begin{enumerate}[label=$\mathrm{R_{cl}}$.,ref=$\mathrm{R_{cl}}$,leftmargin=*]
\item If the things in $A\subseteq\mathcal{T}$ are desirable, then so are the things in $\mathrm{cl}(A)$.\label{rulecl}
\end{enumerate}

I invite the reader to observe that for an operator that has this interpretation, the properties in Definition~\ref{def:closop} make perfect sense. In particular, \ref{ax:clos:empty} imposes that an empty assessment does not lead to meaningfull inferences.\footnote{If we were to drop axiom~\ref{ax:clos:empty}, the theory that we are about to develop would still work~\cite{cooman2023:things:logic:arxiv}, provided that axiom~\ref{ax:sodos:cl} further on is imposed to $\mathcal{A}=\emptyset$ as well. We find this less intuitive though. It furthermore would not yield a more expressive theory, since drawing inferences from nothing---like from the empty set---is already build into the theory as it is: that is exactly what axioms~\ref{ax:desirs:des} and~\ref{ax:sodos:des} are for. We do not consider this to be a form of inference though, and therefore prefer to separate this from the axioms involving the closure operator $\closure$, by requiring that $\closure(\emptyset)=\emptyset$.}

\begin{example}\label{ex:pizzaclosure}
Going back to our pizza example once more, a possible inference mechanism could be that if pizza cheese is deemed desirable, then every desirable pizza will continue to be desirable if we replace its crust with a cheese crust. This corresponds to the closure operator $\closure$ that, for every set $A\subseteq\things$ of pizzas returns the closure
\begin{equation*}
\closure(A)\coloneqq A\cup\{\mathrm{pizza~with~cheese~crust}\colon \mathrm{pizza}\in A\}
\end{equation*}
if $\mathrm{cheese}\in A$ and $\closure(A)\coloneqq A$ otherwise.
The closure of $A=\{\mathrm{meatballs}, \mathrm{cheese}\}$ would then be $\closure(A)=A\cup\{\mathrm{meatballs~with~cheese~crust,~cheese~with~cheese~crust}\}$.
\end{example}

\begin{example}\label{example:preferencesasthings}
For another example, let $\mathcal{O}$ be some set of options, let $\mathcal{T}=\mathcal{T}_\mathcal{O}\coloneqq\{(o_1,o_2)\colon o_1,o_2\in\mathcal{O}\}$ be the set of all ordered pairs of such options, and call a pair $(o_1,o_2)\in\mathcal{T}$ desirable if our subject prefers $o_1$ over $o_2$. In that context, one might want to impose transitivity. To do so, we can let $\closure$ to be the operator $\trans$ that, for every set of preferences $A\subseteq\mathcal{T}$, returns the closure
\begin{equation}\label{eq:transitivity}
\trans(A)
\coloneqq\{(o_0,o_n)\colon o_i\in\mathcal{O}\text{ for all }0\leq i\leq n, (o_{i-1},o_i)\in A\text{ for all }1\leq i\leq n\}
\end{equation}
 of $A$ with respect to transitivity. So $\trans(A)$ contains the preferences that can be inferred from the ones in $A$ by applying transitivity. For this particular choice of closure operator $\closure$, the closure of $A=\{(o_1, o_2),(o_2, o_3)\}$, with $o_1,o_2,o_3$ all different, would then for example be $\closure(A)=\trans(A)=\{(o_1,o_2),(o_2,o_3),(o_1,o_3)\}$. %We will study this choice of $\mathcal{T}$ and $\closure$ in much more detail in Section~\ref{sec:preferences}.
\end{example}

We will call a model coherent if it adheres to the rules \ref{rulenot}, \ref{ruledes} and~\ref{rulecl}---and, in the case of sets of desirable sets, if it is furthermore compatible with our interpretation for a desirable set. For sets of desirable things, this is straightforward and self-explanatory.

\begin{definition}
\label{def:cohdesir}
A set of desirable things $D$ is \emph{coherent} if
\begin{enumerate}[label=$\mathrm{D}_{\arabic*}$.,ref=$\mathrm{D}_{\arabic*}$,leftmargin=*]
\item\label{ax:desirs:not} $A_\mathrm{not}\cap\desirset=\emptyset$
\item\label{ax:desirs:des} $A_\mathrm{des}\subseteq D$
\item\label{ax:desirs:cl} $\closure(\desirset)=\desirset$
\end{enumerate}
We denote the set of all coherent sets of desirable things by $\desirsets$.
\end{definition}

%Note that in the important case where $\mathcal{A}_\mathrm{des}=\{\{t\}\colon t\in A_\mathrm{des}\}$,~\ref{ax:desirs:des} reduces to the requirement that $A_\mathrm{des}\subseteq D$.

For sets of desirable sets, the definition of coherence requires a new piece of machinery: for any subset $\mathcal{A}$ of $\mathcal{P}(\mathcal{T})$, we let 
\begin{equation*}
\mathcal{S}_\mathcal{A}\coloneqq
\big\{
\{t_A\colon A\in\mathcal{A}\}\colon t_A\in A\text{ for all }A\in\mathcal{A}
\big\}
\end{equation*}\label{ex:pizza:selections}
be the set of all selections from $\mathcal{A}$. Any such selection $S\in\mathcal{S}_\mathcal{A}$ is a set that is obtained by selecting from each $A\in\mathcal{A}$ a single thing $t_A$; that is, $S=\{t_A\colon A\in\mathcal{A}\}$. The set $\mathcal{S}_\mathcal{A}$ is simply the set of all sets $S$ that can be obtained in this way.
\begin{example}
Going back to the pizza example, let $\mathcal{A}=\{A_1,A_2\}$, with $A_1=\{\mathrm{peperoni},\mathrm{meatballs}\}$ and $A_2=\{\mathrm{peperoni},\mathrm{cheese}\}$. In that case, $t_{A_1}\in A_1$ can be either peperoni or meatballs, whereas $t_{A_2}\in A_2$ can be peperoni or cheese. Each of the four possible combinations corresponds to a selection, so we find that the set of selections that corresponds to $\mathcal{A}$ is
\begin{equation*}
\mathcal{S}_\mathcal{A}=\big\{
\{\mathrm{peperoni}\},
\{\mathrm{peperoni},\mathrm{cheese}\},
\{\mathrm{meatballs},\mathrm{peperoni}\},
\{\mathrm{meatballs},\mathrm{cheese}\}
\big\}.
\end{equation*}
\end{example}

Using this notion of selections, we now define coherence for sets of desirable sets as follows.

\begin{definition}
\label{def:cohsodos}
A set of desirable sets (of things) $K$ is \emph{coherent} if
\begin{enumerate}[label=$\mathrm{K}_{\arabic*}$.,ref=$\mathrm{K}_{\arabic*}$,leftmargin=*]
\item\label{ax:sodos:nonempty} $\emptyset\notin K$
\item\label{ax:sodos:mon} if $A\subseteq B$ and $A\in K$, then also $B\in K$
\item\label{ax:sodos:not} if $A\in K$ then also $A\setminus A_\mathrm{not}\in K$
\item\label{ax:sodos:des} $\{t\}\in K$ for all $t\in A_\mathrm{des}$
\item\label{ax:sodos:cl} if $\emptyset\neq\mathcal{A}\subseteq K$ and, for all $S\in\mathcal{S}_\mathcal{A}$, $t_S\in\closure(S)$, then $\{t_S\colon S\in\mathcal{S}_\mathcal{A}\}\in K$
\end{enumerate}
We denote the set of all coherent sets of desirable sets (of things) by $\sodoss$.
\end{definition}

The first two of these five axioms are straightforward consequences of the fact that $K$ consists of desirable sets. Indeed, since a set is desirable if it contains at least one desirable thing, it follows at once that the empty set cannot be desirable (\ref{ax:sodos:nonempty}) and that supersets of desirable sets should be desirable as well (\ref{ax:sodos:mon}). 

The last three of these five axioms implement the rules~\ref{rulenot}, \ref{ruledes} and~\ref{rulecl}, again taking into account that $K$ consists of desirable sets.
The third axiom (\ref{ax:sodos:not}) implements \ref{rulenot}, which says that the things in $A_\mathrm{not}$ cannot be desirable. If a set $A$ contains at least one desirable thing, this clearly implies that $A\setminus A_\mathrm{not}$ should also contain at least one desirable thing.
The fourth axiom (\ref{ax:sodos:des}) implements \ref{ruledes}, which simply states that every thing in $A_\mathrm{des}$ should be desirable.
The fifth axiom (\ref{ax:sodos:cl}), finally, implements \ref{rulecl}, which says that if all the things in a set $A$ are desirable then the same is true for $\closure(A)$. To see how this indeed leads to~\ref{ax:sodos:cl}, the crucial observation is that $\mathcal{S}_\mathcal{A}$ contains at least one selection $S$ whose elements are all desirable. The reason is that every $A\in\mathcal{A}\subseteq K$ contains at least one desirable thing. So if, for each $A\in\mathcal{A}$, we let $t_A\in A$ be that desirable thing, then $S=\{t_A\colon A\in\mathcal{A}\}\in\mathcal{S}_\mathcal{A}$ consists of desirable things only. For that particular $S\in\mathcal{S}_\mathcal{A}$, \ref{rulecl} therefore imposes that $\closure(S)$ consists of desirable things only, and thus in particular, that $t_S\in\closure(S)$ is desirable. It follows that $\{t_S\colon S\in\mathcal{S}_\mathcal{A}\}$ contains at least one desirable thing, making it a desirable set. For that reason, it makes sense to require that it belongs to $K$, as~\ref{ax:sodos:cl} does.

\begin{example}\label{ex:pizzaclosure}
To illustrate~\ref{ax:sodos:cl}, consider the closure operator $\closure$ of Example~\ref{ex:pizzaclosure} and the assessment $\mathcal{A}=\{A_1,A_2\}$ of Example~\ref{ex:pizza:selections}. 
 For each of the three $S\in\mathcal{S}_\mathcal{A}$ that contain $\mathrm{peperoni}$, let $t_S\coloneqq\mathrm{peperoni}\in S\subseteq\closure{S}$. For $S=\{\mathrm{meatballs},\mathrm{cheese}\}$, let $t_S\coloneqq\mathrm{meatballs~with~cheese~crust}\in\closure(S)$. If $A_1$ and $A_2$ both belong to $K$, it then follows from~\ref{ax:sodos:cl} that 
\begin{equation*}
\{t_S\colon S\in\mathcal{S}_\mathcal{A}\}
=\{\mathrm{peperoni},\mathrm{meatballs~with~cheese~crust}\}
\end{equation*}
belongs to $K$ as well. To understand why this is reasonable, recall that $A_1,A_2\in K$ means that $A_1$ and $A_2$ are both desirable: they each contain at least one desirable thing. In case $\mathrm{peperoni}$ is not desirable, this implies that $\mathrm{meatballs}$ and $\mathrm{cheese}$ are both desirable, and therefore also, by applying $\closure$, that $\mathrm{meatballs~with~cheese~crust}$ is desirable. So we indeed find that $\mathrm{peperoni}$ or $\mathrm{meatballs~with~cheese~crust}$ should be desirable.
\end{example}

An important feature of the rules and axioms above is that they can accommodate a wide range of different context, by considering various sets of things $\things$ and various choices of $A_\mathrm{not}$, $A_\mathrm{des}$ and $\closure$. The examples above served as a first simple illustration of this range; more complex examples will be given further on. It is also not necessary to impose all of the rules; each of them is optional. Not imposing~\ref{rulenot} or \ref{ruledes} corresponds to setting $A_\mathrm{not}=\emptyset$ and $A_\mathrm{des}=\emptyset$, respectively, whereas not imposing \ref{rulecl} corresponds to letting $\closure$ be equal to the identity operator $\iden$, defined for each $A\subseteq\things$ by $\iden(A)\coloneqq A$. That the corresponding axioms are in those cases indeed redundant is easy to see, except for axiom~\ref{ax:sodos:cl}, for which it is not obvious that it has no implications in case $\closure=\iden$. Nevertheless, the following result shows that it indeed does not.

\begin{proposition}\label{prop:identityoperatorhasnoeffect}
If $\closure=\iden$, then any set of desirable sets $K\subseteq\setsofthings$ that satisfies~\ref{ax:sodos:mon} will also satisfy~\ref{ax:sodos:cl}.
\end{proposition}
\begin{lemma}\label{lem:trickforidentityoperator}
Consider any non-empty $\mathcal{A}\subseteq\setsofthings$ and, for all $S\in\mathcal{S}_\mathcal{A}$, some $t_S\in S$. Then there is some $A\in\mathcal{A}$ such that $A\subseteq\{t_S\colon S\in\mathcal{S}_\mathcal{A}\}$.
\end{lemma}
\begin{proof}
Let $B\coloneqq\{t_S\colon S\in\mathcal{S}_\mathcal{A}\}$. If $A\setminus B\neq\emptyset$ for all $A\in\mathcal{A}$, then (using the axiom of choice) we can choose some $t^*_A\in A\setminus B$ for all $A\in\mathcal{A}$, resulting in a selection $S^*\coloneqq\{t^*_A\colon A\in\mathcal{A}\}\in\mathcal{S}_\mathcal{A}$ such that $S^*\cap B=\emptyset$. This is however impossible because $t_{S^*}\in S^*$ and $t_{S^*}\in B$. So we see that there must be some $A\in\mathcal{A}$ such that $A\setminus B=\emptyset$, or equivalently, $A\subseteq B$. 
\end{proof}
\begin{proof}[Proof of Proposition~\ref{prop:identityoperatorhasnoeffect}]
To prove \ref{ax:sodos:cl}, consider any $\emptyset\neq\mathcal{A}\subseteq K$ and, for all $S\in\mathcal{S}_\mathcal{A}$, some $t_S\in\closure(S)$, and let $B\coloneqq\{t_S\colon S\in\mathcal{S}_\mathcal{A}\}$. We need to prove that $B\in K$. For all $S\in\mathcal{S}_\mathcal{A}$, since $\closure=\iden$, it follows from $t_S\in\closure(S)$ that $t_S\in S$. It therefore follows from Lemma~\ref{lem:trickforidentityoperator} that there is some $A\in\mathcal{A}$ such that $A\subseteq B$. Since $A\in\mathcal{A}\subseteq K$, it therefore follows from \ref{ax:sodos:mon} that $B\in K$.
\end{proof}

We end this section by addressing the question whether it is at all possible to be coherent; that is, whether the proposed axioms are compatible. The following result presents a simple intuitive condition that is both necessary and sufficient for this to be the case.

\begin{proposition}\label{prop:coherenceispossible}
$\desirsets$ is non-empty---so there is at least one coherent set of desirable things---if and only if $\closure(A_\mathrm{des})\cap A_\mathrm{not}=\emptyset$.
Similarly, $\sodoss$ is non-empty---so there is at least one coherent set of desirable sets---if and only if $\closure(A_\mathrm{des})\cap A_\mathrm{not}=\emptyset$.
\end{proposition}
\begin{proof}
We provide a circular proof, thus showing that the three conditions in the statement are all equivalent.

First assume that $\closure(A_\mathrm{des})\cap A_\mathrm{not}=\emptyset$, and let $D\coloneqq\closure(A_\mathrm{des})$. It then follows directly from~\ref{ax:clos:ext} and~\ref{ax:clos:idem} that $D$ is coherent, which implies that $\desirsets$ is non-empty. 

Next, assume that $\desirsets$ is non-empty, meaning that there is at least one coherent set of desirable things $\desirset$. It then follows from Proposition~\ref{prop:DcoherentiffKDis} further on that $K_D$---see Section~\ref{sec:representation}---is a coherent set of desirable sets, so $\sodoss$ is then non-empty too.

Finally, assume that $\sodoss$ is non-empty, meaning that there is at least one coherent set of desirable sets $K$. We set out to prove that $\closure(A_\mathrm{des})\cap A_\mathrm{not}=\emptyset$. If $A_\mathrm{des}=\emptyset$, this follows trivially from~\ref{ax:clos:empty}. To show that it is also true if $A_\mathrm{des}\neq\emptyset$, we let $\mathcal{A}^*\coloneqq\{\{t\}\colon t\in A_\mathrm{des}\}\neq\emptyset$. Then $\mathcal{S}_{\mathcal{A}^*}=\{A_\mathrm{des}\}$ and, due to \ref{ax:sodos:des}, $\mathcal{A^*}\subseteq K$. For any $t\in\closure(A_\mathrm{des})$, it therefore follows from~\ref{ax:sodos:cl} that $\{t\}\in K$. Now assume \emph{ex absurdo} that $\closure(A_\mathrm{des})\cap A_\mathrm{not}\neq\emptyset$ and consider any $t$ in this intersection. Then since $t\in\closure(A_\mathrm{des})$, as argued above, $\{t\}\in K$. Since also $t\in A_\mathrm{not}$, it therefore follows from~\ref{ax:sodos:not} that $\emptyset\in K$, contradicting~\ref{ax:sodos:nonempty}. Hence, in all cases, $\closure(A_\mathrm{des})\cap A_\mathrm{not}=\emptyset$.
\end{proof}

\section{Representation}\label{sec:representation}

An essential feature of the theory of desirable things here presented, is the connection between the two frameworks it consists of. We start by showing that sets of desirable sets (of things) are more expressive than sets of desirable things. The reason, quite simply, is that every set of desirable things $\desirset$ has a corresponding set of desirable sets
\begin{equation}\label{eq:desirset:to:rejectset}
\rejectset[\desirset]
\coloneqq\cset{\optset\in\mathcal{P}(\mathcal{T})}{\optset\cap\desirset\neq\emptyset}.
\end{equation}
Furthermore, one of them is coherent if and only if the other is.
\begin{proposition}\label{prop:DcoherentiffKDis}
%\textcolor{orange}{
Consider a set of desirable things $\desirset$ and its corresponding set of desirable sets $\sodos[\desirset]$. Then $\desirset$ is coherent if and only if $\sodos[\desirset]$ is.%}
\end{proposition}
\begin{proof}%{\textit{\textbf{of Proposition~\ref{prop:DcoherentiffKDis} }}}
Consider any set of desirable things $\desirset$ and let $\rejectset[\desirset]$ be its corresponding set of desirable sets.

For the `only if'-part of the statement, we assume that $\desirset$ is coherent and prove that $\sodos[\desirset]$ is then coherent as well. That $\sodos[\desirset]$ satisfies~\ref{ax:sodos:nonempty} and~\ref{ax:sodos:mon} follows directly from Equation~\eqref{eq:desirset:to:rejectset}. To see that $\sodos[\desirset]$ satisfies \ref{ax:sodos:not}, observe that for any $A\in\sodos[\desirset]$, it follows from Equation~\eqref{eq:desirset:to:rejectset} that there is some $t\in A\cap\desirset$. Due to~\ref{ax:desirs:not}, we furthermore know that $t\notin A
_{\mathrm{not}}$. Hence, $t\in A\setminus A_{\mathrm{not}}$. Since $t\in\desirset$, this implies that $A\setminus A_{\mathrm{not}}\cap\desirset\neq\emptyset$.
To see that $\rejectset[\desirset]$ satisfies~\ref{ax:sodos:des}, observe that for any $t\in A_\mathrm{des}$, \ref{ax:desirs:des} implies that $t\in D$ and therefore, that $\{t\}\cap D\neq\emptyset$ and hence $\{t\}\in K_D$.
To see that $\rejectset[\desirset]$ also satisfies~\ref{ax:sodos:cl}, we consider any $\emptyset\neq\assessment\subseteq\sodos[\desirset]$ and, for all $S\in\mathcal{S}_\mathcal{A}$, some $t_S\in\closure(S)$. For all $A\in\mathcal{A}$, since $A\in\sodos[\desirset]$, there is some $t^*_A\in A\cap\desirset$. Hence, if we let $S^*\coloneqq\{t^*_A\colon A\in\mathcal{A}\}\in\mathcal{S}_\mathcal{A}$, then $S^*\subseteq\desirset$ and therefore, due to~\ref{ax:clos:mon} and~\ref{ax:desirs:cl}, $\closure(S^*)\subseteq\closure(\desirset)=\desirset$. This implies that $t_{S^*}\in\desirset$ and therefore, since $t_{S^*}\in\{t_S\colon S\in\mathcal{S}_\mathcal{A}\}$, that $\{t_S\colon S\in\mathcal{S}_\mathcal{A}\}\in\sodos[\desirset]$.

For the `if'-part of the statement, we assume that $\sodos[\desirset]$ is coherent and prove that $\desirset$ is then coherent as well. To prove that $\desirset$ satisfies~\ref{ax:desirs:not}, we assume \emph{ex absurdo} that $A_\mathrm{not}\cap\desirset\neq\emptyset$. Consider any $t\in A_\mathrm{not}\cap\desirset$. 
Then since $\{t\}\cap\desirset\neq\emptyset$, we have that $\{t\}\in\sodos[\desirset]$. 
Due to \ref{ax:sodos:not}, this implies that $\emptyset=\{t\}\setminus A_\mathrm{not}\in\rejectset$, contradicting~\ref{ax:sodos:nonempty}. 
To prove that $\desirset$ satisfies~\ref{ax:desirs:des}, consider any $t\in A_\mathrm{des}$. Since $\{t\}\in K_D$ because of~\ref{ax:sodos:des}, it then follows that $\{t\}\cap D\neq\emptyset$ and therefore $t\in D$.
To prove that $\desirset$ satisfies~\ref{ax:desirs:cl}, we consider any $t^*\in\closure(\desirset)$ and prove that $t^*\in\desirset$; \ref{ax:desirs:cl} then follows from~\ref{ax:clos:ext}. 
Let $\mathcal{A}\coloneqq\{\{t\}\colon t\in\desirset\}\subseteq\sodos[\desirset]$. 
Then $\mathcal{A}\neq\emptyset$ because $t^*\in\closure(D)$ implies that $\closure(D)\neq\emptyset$ and therefore, using~\ref{ax:clos:empty}, that $D\neq\emptyset$.
Furthermore, we also clearly have that $\mathcal{S}_\mathcal{A}=\{\desirset\}$. 
Since $t^*\in\closure(\desirset)$, it therefore follows from~\ref{ax:sodos:cl} that $\{t^*\}\in\sodos[\desirset]$, or equivalently, that $t^*\in\desirset$.
\end{proof}

More involved sets of desirable sets can for example be obtained by taking intersections of these basic ones. In particular, with any non-empty set $\setofdesirsets$ of sets of desirable things, we can associate a set of desirable sets
\begin{equation*}
\sodos[\setofdesirsets]\coloneqq\bigcap_{\desirset\in\setofdesirsets}\sodos[\desirset].
\end{equation*}
This kind of model expresses that at least one of the SDTs in $\setofdesirsets$ is representative, in the sense that a set of things is deemed desirable according to $K_\setofdesirsets$ if and only if it is desirable according to all $D\in\setofdesirsets$. If every set of desirable things in $\setofdesirsets$ is coherent, the resulting set of desirable sets $K_\setofdesirsets$ will furthermore be coherent as well. This follows from Proposition~\ref{prop:DcoherentiffKDis} and the fact that coherence is preserved under intersections.
\begin{proposition}\label{prop:intersectionofcoherentKsiscoherent}
For every non-empty set $\mathcal{K}\subseteq\sodoss$ of coherent sets of desirable sets, $\bigcap\mathcal{K}$ is a coherent set of desirable sets as well. Similarly, for every non-empty set $\setofdesirsets\subseteq\desirsets$ of coherent sets of desirable things, $\bigcap\setofdesirsets$ is a coherent set of desirable things.
\end{proposition}
\begin{proof}Every $K\in\mathcal{K}$ is coherent, and therefore satisfies~\ref{ax:sodos:nonempty}--\ref{ax:sodos:cl}. Since these five axioms are preserved under arbitrary intersections, and since $\mathcal{K}$ is non-empty, it follows that $\bigcap\mathcal{K}$ satisfies~\ref{ax:sodos:nonempty}--\ref{ax:sodos:cl} as well, and therefore, that it is coherent.

The proof for $\setofdesirsets$ is equivalent. The only aspect that is perhaps not trivial is that \ref{ax:desirs:cl} is preserved under intersections. To see this, consider any two sets of desirable things $D_1,D_2$ that satisfy \ref{ax:desirs:cl} and let $D\coloneqq D_1\cap D_2$. Then 
\begin{equation*}
\closure(D)=\closure(D_1\cap D_2)\subseteq\closure(D_1)=D_1,
\end{equation*}
using~\ref{ax:clos:mon} for the inclusion and \ref{ax:desirs:cl} for the second equality. Since we similarly have that $\closure(D)\subseteq D_2$, it follows that $\closure(D)\subseteq D_1\cap D_2=D$. A final application of \ref{ax:clos:mon} therefore yields $\closure(D)=D$.
\end{proof}
\begin{corollary}\label{corol:fromsetofDtoK}
For any non-empty set $\setofdesirsets\subseteq\desirsets$ of coherent sets of desirable things, $K_\setofdesirsets$ is a coherent set of desirable sets.
\end{corollary}
\begin{proof}
Immediate consequence of Propositions~\ref{prop:DcoherentiffKDis} and~\ref{prop:intersectionofcoherentKsiscoherent}. 
\end{proof}

What is much more surprising, and a central result of this report, is that the converse is true as well: every coherent set of desirable sets $\sodos$ corresponds to a set $\setofdesirsets$ of coherent sets of desirable things. One such set will be of the form
\begin{equation}\label{eq:representingsetfromA}
\setofdesirsets_\mathcal{A}\coloneqq\{
  \closure(S\cup A_\mathrm{des})\colon S\in\mathcal{S}_\mathcal{A}, \closure(S\cup A_\mathrm{des})\cap A_\mathrm{not}=\emptyset
\},
\end{equation}
where $\mathcal{A}\subseteq\mathcal{P}(\mathcal{T})$ and, for this result in particular, $\mathcal{A}=K$.

\begin{theorem}\label{theo:infinite:coherentrepresentation:twosided}
A set of desirable sets (of things) $\sodos$ is coherent if and only if there is a non-empty set $\setofdesirsets\subseteq\desirsets$ of coherent sets of desirable things such that $\sodos=\sodos[\setofdesirsets]$. One such set $\setofdesirsets$ is $\mathcal{D}_K$, and the largest such set $\setofdesirsets$ is $\desirsets(\sodos)\coloneqq\cset{\desirset\in\desirsets}{\sodos\subseteq\sodos[\desirset]}$.
\end{theorem}

Most of the heavy lifting in the proof is done by Lemma~\ref{lem:DAproperties}, which itself makes use of the following strengthened version of~\ref{ax:sodos:cl} that drops the condition that $\mathcal{A}\neq\emptyset$ and otherwise essentially combines~\ref{ax:sodos:cl} with~\ref{ax:sodos:des}.

\begin{lemma}\label{lem:combinedesandcl}
Let $K$ be a coherent set of desirable sets. If $\mathcal{A}\subseteq K$ and, for all $S\in\mathcal{S}_\mathcal{A}$, $t_S\in\closure(S\cup A_\mathrm{des})$, then $\{t_S\colon S\in\mathcal{S}_\mathcal{A}\}\in K$
\end{lemma}
\begin{proof}
Let $B\coloneqq\{t_S\colon S\in\mathcal{S}_\mathcal{A}\}$. We set out to prove that $B\in K$.
To that end, let $\mathcal{A}^*\coloneqq\mathcal{A}\cup\{\{t\}\colon t\in A_\mathrm{des}\}$.
We distinguish two cases: $\mathcal{A}\neq\emptyset$ and $\mathcal{A}=\emptyset$. If $\mathcal{A}\neq\emptyset$, then clearly also $\mathcal{A}^*\neq\emptyset$. If $\mathcal{A}=\emptyset$, then $\mathcal{S}_\mathcal{A}=\{\emptyset\}$ and therefore $B=\{t_\emptyset\}$, with $t_\emptyset\in\closure(A_\mathrm{des})$. This implies that $\closure(A_\mathrm{des})\neq\emptyset$, and therefore also $A_\mathrm{des}\neq\emptyset$ because of~\ref{ax:clos:empty}, which in turn implies that $\mathcal{A}^*\neq\emptyset$. Hence, in both cases, $\mathcal{A^*}\neq\emptyset$. Furthermore, we also have that $\mathcal{A}^*\subseteq K$ because $\mathcal{A}\subseteq K$ and $K$ satisfies~\ref{ax:sodos:des}.

 Consider now any $S^*\in\mathcal{S}_{\mathcal{A}^*}$. Then on the one hand, since $\mathcal{A}\subseteq\mathcal{A}^*$, there is some $S\in\mathcal{S}_\mathcal{A}$ such that $S\subseteq S^*$. On the other hand, since $\{\{t\}\colon t\in A_\mathrm{des}\}\subseteq\mathcal{A}^*$, we also know that $A_\mathrm{des}\subseteq S^*$. So $S\cup A_\mathrm{des}\subseteq S^*$. Hence, if we let $t_{S^*}\coloneqq t_S$, we find that $t_{S^*}=t_S\in\closure(S\cup A_\mathrm{des})\subseteq\closure(S^*)$, using~\ref{ax:clos:mon} for the last inclusion, and that $t_{S^*}=t_S\in B$. Selecting such a $t_{S^*}$ for each $S^*\in\mathcal{S}_{\mathcal{A}^*}$, we arrive at a set $B^*\coloneqq\{t_{S^*}\colon S^*\in\mathcal{S}_{\mathcal{A}^*}\}\subseteq B$ such that, for each $S^*\in\mathcal{S}_{\mathcal{A}^*}$, $t_{S^*}\in\closure(S^*)$. Since $\emptyset\neq\mathcal{A}^*\subseteq K$, \ref{ax:sodos:cl} implies that $B^*\in K$. Since $B^*\subseteq B$, \ref{ax:sodos:mon} therefore implies that $B\in K$.
\end{proof}

\begin{lemma}\label{lem:DAproperties}
Let $\sodos$ be a coherent SDS. Then for any $\mathcal{A}\subseteq K$, $\mathcal{D}_\mathcal{A}$ is a non-empty subset of $\desirsets$ such that $\mathcal{A}\subseteq\sodos[\setofdesirsets_\mathcal{A}]\subseteq\sodos$.
\end{lemma}

\begin{proof}%[Proof of Lemma~\ref{lem:DAproperties}]
To prove that $\setofdesirsets_\mathcal{A}$ is non-empty, 
%we consider two cases: $\mathcal{A}=\emptyset$ and $\mathcal{A}\neq\emptyset$. If $\mathcal{A}=\emptyset$, then $\mathcal{S}_\mathcal{A}=\{\emptyset\}$ and therefore, since the coherence of $K$ and Proposition~\ref{prop:coherenceispossible} imply that $\closure(A_\mathrm{des})\cap A_\mathrm{not}=\emptyset$, $\setofdesirsets_\mathcal{A}=\{\closure(A_\mathrm{des})\}\neq\emptyset$. 
% For the case $\mathcal{A}\neq\emptyset$, 
we assume \emph{ex absurdo} that $\setofdesirsets_\mathcal{A}$ is empty. This then implies that for all $S\in\mathcal{S}_\mathcal{A}$, $\closure(S\cup A_\mathrm{des})\cap A_\mathrm{not}\neq\emptyset$. Hence, for all $S\in\mathcal{S}_\mathcal{A}$, we can choose some $t_S\in\closure(S\cup A_\mathrm{des})\cap A_\mathrm{not}$. Then on the one hand, since $\mathcal{A}\subseteq K$, it follows from Lemma~\ref{lem:combinedesandcl} that $\{t_S\colon S\in\mathcal{S}_\mathcal{A}\}\in K$. On the other hand, however, we also have that $\{t_S\colon S\in\mathcal{S}_\mathcal{A}\}\subseteq A_\mathrm{not}$. It therefore follows from~\ref{ax:sodos:mon} that $A_\mathrm{not}\in K$, which, due to~\ref{ax:sodos:not}, implies that $\emptyset\in K$, contradicting~\ref{ax:sodos:nonempty}. Hence, $\setofdesirsets_\mathcal{A}$ is non-empty.

Consider now any $D\in\setofdesirsets_\mathcal{A}$. It then follows from Equation~\eqref{eq:representingsetfromA} that there is some $S\in\mathcal{S}_\mathcal{A}$ such that $D=\closure(S\cup A_\mathrm{des})$ and $\closure(S\cup A_\mathrm{des})\cap A_{\mathrm{not}}\neq\emptyset$, with $S=\{t_A\colon A\in \mathcal{A}\}$ and, for all $A\in \mathcal{A}$, $t_A\in A$. Since $D=\closure(S\cup A_\mathrm{des})$ and $\closure(S\cup A_\mathrm{des})\cap A_{\mathrm{not}}\neq\emptyset$, we immediately see that $D$ satisfies~\ref{ax:desirs:not}. That it also satisfies~\ref{ax:desirs:des} follows from~\ref{ax:clos:ext}, since that tells us that $A_\mathrm{des}\subseteq S\cup A_\mathrm{des}\subseteq\closure(S\cup A_\mathrm{des})=D$, and that it satisfies~\ref{ax:desirs:cl} follows from~\ref{ax:clos:idem}, since that tells us that $\closure(D)=\closure(\closure(S\cup A_\mathrm{des}))=\closure(S\cup A_\mathrm{des})=D$. Hence, $D\in\desirsets$. Furthermore, for any $A\in\mathcal{A}$, since $t_A\in S$, it follows from~\ref{ax:clos:ext} that $t_A\in S\cup A_\mathrm{des}\subseteq\closure(S\cup A_\mathrm{des})=\desirset$, which, since $t_A\in A$, implies that $A\cap\desirset\neq\emptyset$ and, therefore, that $A\in\sodos[\desirset]$. Since this is true for any $A\in\mathcal{A}$, it follows that $\mathcal{A}\subseteq\sodos[\desirset]$. So in summary, for any $\desirset\in\setofdesirsets_\mathcal{A}$, we have found that $\desirset\in\desirsets$ and $\mathcal{A}\subseteq\sodos[\desirset]$. Hence, $\setofdesirsets_\mathcal{A}\subseteq\desirsets$ and $\mathcal{A}\subseteq\cap_{\desirset\in\setofdesirsets_\mathcal{A}}\sodos[\desirset]=\sodos[\setofdesirsets_\mathcal{A}]$.

The only thing left to prove is that $\sodos[\setofdesirsets_\mathcal{A}]\subseteq\sodos$. To that end, we consider any $A\in\sodos[\setofdesirsets_\mathcal{A}]$ and set out to prove that $A\in\sodos$. Consider any $S\in\mathcal{S}_\mathcal{A}$. If $\closure(S\cup A_\mathrm{des})\cap A_\mathrm{not}\neq\emptyset$, then choose some $t_S\in\closure(S\cup A_\mathrm{des})\cap A_\mathrm{not}\subseteq\closure(S\cup A_\mathrm{des})\cap(A_\mathrm{not}\cup A)$. If $\closure(S\cup A_\mathrm{des})\cap A_\mathrm{not}=\emptyset$, then $\closure(S\cup A_\mathrm{des})\in\setofdesirsets_\mathcal{A}$ and therefore, $A\in\sodos[\setofdesirsets_\mathcal{A}]=\cap_{\desirset\in\setofdesirsets_\mathcal{A}}\sodos[\desirset]\subseteq\sodos[\closure(S\cup A_\mathrm{des})]$, which implies that $A\cap\closure(S\cup A_\mathrm{des})\neq\emptyset$. In that case, choose some $t_S\in A\cap\closure(S\cup A_\mathrm{des})\subseteq\closure(S\cup A_\mathrm{des})\cap(A_\mathrm{not}\cup A)$. So for all $S\in\mathcal{S}_\mathcal{A}$, we have found some $t_S\in\closure(S\cup A_\mathrm{des})\cap(A_\mathrm{not}\cup A)$.  
Then on the one hand, since $\mathcal{A}\subseteq K$, it follows from Lemma~\ref{lem:combinedesandcl} that $\{t_S\colon S\in\mathcal{S}_\mathcal{A}\}\in K$, which, due to~\ref{ax:sodos:not}, implies that $\{t_S\colon S\in\mathcal{S}_\mathcal{A}\}\setminus A_\mathrm{not}\in K$. On the other hand, however, we also have that $\{t_S\colon S\in\mathcal{S}_\mathcal{A}\}\subseteq A_\mathrm{not}\cup A$, which implies that $\{t_S\colon S\in\mathcal{S}_\mathcal{A}\}\setminus A_\mathrm{not}\subseteq A$. It therefore follows from~\ref{ax:sodos:mon} that $A\in K$, as required.
\end{proof}

\begin{proof}[Proof of Theorem~\ref{theo:infinite:coherentrepresentation:twosided}]
The `if'-part of the statement was already established in Corollary~\ref{corol:fromsetofDtoK}.
For the `only if' part, we assume that $\sodos$ is a coherent set of desirable sets.
We need to prove that there is a non-empty set $\setofdesirsets\subseteq\desirsets$ of coherent sets of desirable things such that $\sodos=\sodos[\setofdesirsets]$, that $\setofdesirsets_K$ provides such a set, and that $\desirsets\group{\sodos}$ is the largest such set. From Lemma~\ref{lem:DAproperties}, with $\mathcal{A}=K$, it follows that $\setofdesirsets_K$ is a non-empty subset of $\desirsets$ and that $\sodos=\sodos[\setofdesirsets_K]$. So $\setofdesirsets_K$ indeed satisfies the properties that are required from $\setofdesirsets$ in our theorem. This therefore also establishes that there is at least one set $\setofdesirsets$ that satisfies these properties. So it remains to prove that $\desirsets\group{\sodos}$ is the largest set that satisfies these properties. That any non-empty set $\mathcal{D}\subseteq\desirsets$ such that $\sodos=\sodos[\setofdesirsets]$ is included in $\desirsets\group{\sodos}$ is immediate because, for any $D\in\mathcal{D}\subseteq\desirsets$, $\sodos=\sodos[\mathcal{D}]\subseteq\sodos[D]$. So we are left to show that $\desirsets\group{\sodos}$ itself satisfies the properties as well. We already know there is a non-empty subset $\mathcal{D}$ of $\desirsets$ such that $\sodos=\sodos[\setofdesirsets]$, and we also know that this implies that $\mathcal{D}$ is included in $\desirsets(\sodos)$. Since $\mathcal{D}$ is non-empty and $\setofdesirsets\subseteq\desirsets(\sodos)$, it follows that $\desirsets(\sodos)$ is non-empty too and, since $\sodos=\sodos[\setofdesirsets]$ and $\setofdesirsets\subseteq\desirsets(\sodos)$, it follows that $\sodos[\desirsets(\sodos)]\subseteq\sodos[\setofdesirsets]=\sodos$. 
By definition, $\desirsets(\sodos)$ is furthermore a subset of $\desirsets$ such that $\sodos\subseteq\sodos[\desirsets(\sodos)]$. So we find that, indeed, $\desirsets(\sodos)$ is a non-empty subset of $\desirsets$, and that $\sodos=\sodos[\desirsets(\sodos)]$.
\end{proof}

As a first serious illustration of the proposed framework, and of Theorem~\ref{theo:infinite:coherentrepresentation:twosided} in particular, we apply it to the case of desirable gambles, leading to variations on existing results for so-called coherent sets of desirable gambles, as well as for generalisations thereof where instead of gambles, arbitrary vectors are considered.

\begin{example}\label{ex:gambles}
Let $\mathcal{X}$ be a set of states, typically regarded as the possible outcomes of some experiment, and let $\mathcal{G}(\mathcal{X})$ be the set of all bounded real functions on $\mathcal{X}$, called gambles. Any gamble $f$ in $\mathcal{G}(\mathcal{X})$ can then be interpreted as an uncertain reward $f(x)$ whose value depends on the unkown state $x$. Several notions of desirability could be considered here, but one that is particulary intuitive and common is to call such a gamble desirable if our subject (strictly) prefers this gamble---so receiving the reward $f(x)$ after the state $x$ has been determined---to the status quo---so to not gambling at all.

The definitions of coherence that are commonly used for sets of desirable gambles then correspond to particular cases of Definition~\ref{def:cohdesir}~\cite{quaeghebeur2012:itip,couso2011:desirable}.
One popular choice, that makes sense if desirability is defined as above, is to let $A_\mathrm{not}\coloneqq\{f\in\mathcal{G}(\mathcal{X})\colon f\leq0\}$ and
\begin{equation}\label{eq:Adesgambles}
A_\mathrm{des}\coloneqq\{f\in\mathcal{G}(\mathcal{X})\colon f\geq0\text{ and }(\exists x\in\mathcal{X})f(x)>0\},
\end{equation}
and to let $\closure\coloneqq\posi$ be the closure operator with respect to positive linear combinations, defined for all $A\subseteq\mathcal{G}(\mathcal{X})$ by
\begin{equation*}
\posi(A)\coloneqq\{\sum_{i=1}^n\lambda_i f_i\colon n>0,f_i\in A,\lambda_i>0\}.
\end{equation*}
In this case, $A_\mathrm{not}$ expresses that a gamble that only yields negative rewards is not desirable, $A_\mathrm{des}$ expresses that a nonnegative but possibly positive reward should always be desirable, and $\closure$ corresponds to an assumption that the awards are expressed in units of some linear utility scale. Alternative versions of this concept of a coherent set of desirable gambles are typically very similar, and also correspond to special case of our framework, but with small differences in the choice of $A_\mathrm{des}$ and $A_\mathrm{not}$; $A_\mathrm{des}\coloneqq\{f\in\mathcal{G}(\mathcal{X})\colon \inf(f)>0\}$ is for example often considered as well.
Regardless of the particular choice, Definition~\ref{def:cohsodos} will lead to a corresponding notion of coherent sets of desirable sets of gambles, and Theorem~\ref{theo:infinite:coherentrepresentation:twosided} shows that the latter can always be represented by a set of coherent sets of desirable gambles. A similar result can for example be found in~\cite{debock2018}, but the desirable sets (of gambles) can only be finite there, and the version of axiom~\ref{ax:sodos:cl} used there, with $\closure=\posi$, is therefore simpler. Such simplifications are also possible in our general framework of desirable things; we will get to that in Section~\ref{sec:finitecase}.

Rather than focus on gambles, completely similar concepts and results can also be obtained for vectors in some arbitrary vector space $\mathcal{V}$. The choice of $A_\mathrm{des}$ and $A_\mathrm{not}$ will then be different, and will depend on the particular vector space and context, but the inference mechanism that is expressed by $\closure=\posi$ then still makes sense (because linear combinations do), and all the results continue to apply. In that context, a version of Theorem~\ref{theo:infinite:coherentrepresentation:twosided} can for example be found in~\cite{pmlr-v103-de-bock19b}, be it again for the case where desirable sets can only be finite. Examples of things in a vector space for which the notion of desirability has proven useful include, for example, desirable polynomials~\cite{decooman2015:coherent:predictive:inference} and desirable matrices~\cite{Benavoli_2016}.
\end{example}

The main determining feature of the cases considered in the example above, is that the inference mechanism consists in taking positive linear combinations. However, this is not always defensible. In the case of desirable gambles, for example, one might want to drop the assumption of linear utility that justifies the use of the $\posi$ operator. The framework of desirable things makes this easy; one can easily use a different closure operator such as, for example, the convex hull.

\begin{example}\label{ex:nonlineargambles}
Let $\things$ be the set $\mathcal{G}(\mathcal{X})$ of all gambles, as in Example~\ref{ex:gambles}. With $A_\mathrm{not}\coloneqq\{f\in\mathcal{G}(\mathcal{X})\colon f\leq0\}$ and $A_\mathrm{des}\coloneqq\{f\in\mathcal{G}(\mathcal{X})\colon f\geq0\text{ and }(\exists x\in\mathcal{X})f(x)>0\}$. Rather than choose $\closure\coloneqq\posi$, however, we can drop the assumption of linear utility and use any other closure operator instead. For coherent sets of desirable gambles, this corresponds to the theory of nonlinear desirability of Miranda and Zaffalon~\cite{miranda2022:nonlinear:desirability,ZaffalonSureThing}, who provide several examples for such closure operators. Another example of such a closure operator, which was essentially already put forward by Quaeghebeur~\cite{quaeghebeur2014FUR}, is to let $\closure\coloneqq\chull$ be the convex hull operator, defined for all $A\subseteq\mathcal{G}(\mathcal{X})$ by
\begin{equation*}
\chull(A)\coloneqq\{\sum_{i=1}^n\lambda_i f_i\colon n>0,f_i\in A,\lambda_i>0,\sum_{i=1}^n\lambda_i=1\},
\end{equation*}
which can easily be seen to be a closure operator. In any case, regardless of the particular choice of $\closure$, it follows from our results that it not only leads to a notion of coherent sets of desirable gambles, as in \cite{miranda2022:nonlinear:desirability}, but also to a notion of coherent sets of desirable sets of gambles, as well as a connection between both as provided by Theorem~\ref{theo:infinite:coherentrepresentation:twosided}.
\end{example}

The framework of desirable things is also not restricted to the vector spaces---such as sets of gambles---to which it has typically been applied in the past: moving away from the $\posi$ operator also opens up the possibility of moving away from vector spaces. The only real restriction on $\things$ is that $\closure$ maps subsets of $\things$ to subsets of $\things$. For example, if inference is expressed by means of the convex hull operator, $\things$ can be any convex space. Interestingly, this makes it possible to directly deal with probabilities and horse lotteries. In contrast, previous attempts at applying the ideas behind desirability to horse lotteries consisted in embedding horse lotteries in a vector space and working with positive linear rather than convex combinations in that vector space~\cite{zaffalon2017:incomplete:preferences,2017vancamp:phdthesis,DECOOMAN2022255}.

\begin{example}\label{ex:lotteriesandsuch}
Let $\things$ be the set of all probability mass functions on some finite set $\mathcal{R}$ that consists of `prizes'. Any such mass function is called a lottery, and provides a probability for winning each of the prizes in $\mathcal{R}$. We could then call a lottery desirable if our subject is willing to let this lottery determine which prize he will get from $\mathcal{R}$. The set of things $\things$ is now a convex space, but not a vector space. Nevertheless, the theory we developed can still be applied. Here too, we can for example consider $\chull$ as our closure operator, which builds in an assumption that convex mixtures of desirable lotteries should be desirable as well. $A_\mathrm{not}$ and $A_\mathrm{des}$ can be chosen freely, but will typically depend on the particular set of prizes $\mathcal{R}$. If $\mathcal{R}$ contains a prize that is clearly positive---free dinner with friends, say---it for example makes sense for $A_\mathrm{des}$ to include the lottery that yields $r$ with probability one, whereas if $\mathcal{R}$ contains a `prize' that is clearly negative---no warm meals for a weak, say---the lottery that assigns probability one to that prize would reasonably be included in $A_\mathrm{not}$. For any given such set-up, the theory we developed provides a notion of sets of desirable lotteries, of sets of desirable sets of lotteries, as well as a connection between both types of models.

The example of lotteries on a finite set $\mathcal{R}$ is furthermore completely arbitrary; we can consider any other convex space instead. $\mathcal{R}$ could for example be infinite, and $\things$ could then be the set of all finitely additive probability measures on $\mathcal{P}(\mathcal{R})$, or if we endow $\mathcal{R}$ with a sigma algebra, the set of all countably additive probability measures on that algebra. As another example, $\things$ could be the set of all so-called horse loteries~\cite{anscombe1963} on a state space $\mathcal{X}$ with a finite set of prizes $\mathcal{R}$:
\begin{equation}\label{eq:horselotteries}
\things=\{h\in\mathbb{R}_{\geq0}^{\statesandrewards}\colon \sum_{r\in\rewards} h(x,r)=1\text{ for all }x\in\states\}.
\end{equation}
These horse lotteries are typically interpreted as gambles for which the reward associated with every $x\in\mathcal{X}$ is not numeric, but rather a lottery on $\mathcal{R}$. Here too, $\mathcal{R}$ could also be infinite; the lottery associated with each $x\in\states$ will then be a (finitely or countably additve) probability measure on $\mathcal{R}$ instead of a probability mass function.
\end{example}

\section{Finite desirable sets and simpler axioms}
\label{sec:finitecase}

For sets of desirable sets, the axiom that imposes the effects of $\closure$ is quite demanding: it requires us to combine the effects of infinitely many desirable sets in $\mathcal{A}$ to draw conclusions about the desirability of a new set $\{t_S\colon S\in\mathcal{S}_\mathcal{A}\}$. In contrast, in previous studies of sets of desirable sets of \emph{gambles}~\cite{debock2018,pmlr-v103-de-bock19b}, the axiom that imposes the effects of $\posi$ was simpler than \ref{ax:sodos:cl}, and essentially required~\ref{ax:sodos:cl} only for the case $\abs{\mathcal{A}}=2$. We now intend to investigate whether---and if yes, under which conditions---such simplifications can also be obtained in our more general context, where the things in $\things$ are not necessarily gambles and the considered closure operator is not necessarily the $\posi$ operator. 
We will in particular consider three possible simplifications for \ref{ax:sodos:cl}. First, a version that imposes \ref{ax:sodos:cl} for finite $\mathcal{A}$ only:
\begin{enumerate}[label=$\mathrm{K}_{5\mathrm{fin}}$.,ref=$\mathrm{K}_{5\mathrm{fin}}$,leftmargin=*]
\item\label{ax:sodos:clfin} if $\emptyset\neq\mathcal{A}\subseteq K$ is finite and, for all $S\in\mathcal{S}_\mathcal{A}$, $t_S\in\closure(S)$, \\then $\{t_S\colon S\in\mathcal{S}_\mathcal{A}\}\in K$.
\end{enumerate} 
Second, mimicking the approach in earlier studies for the case of desirable gambles, a version that essentially focusses on the case $\abs{\mathcal{A}}=2$:
\begin{enumerate}[label=$\mathrm{K}_{5\mathrm{bin}}$.,ref=$\mathrm{K}_{5\mathrm{bin}}$,leftmargin=*]
\item\label{ax:sodos:clbin} if $A,B\in K$ and, for all $a\in A$ and $b\in B$, $t_{a,b}\in\closure(\{a,b\})$, \\then $\{t_{a,b}\colon a\in A, b\in B\}\in K$.
\end{enumerate}
And finally, a version that only considers the case $\abs{\mathcal{A}}=1$:
\begin{enumerate}[label=$\mathrm{K}_{5\mathrm{un}}$.,ref=$\mathrm{K}_{5\mathrm{un}}$,leftmargin=*]
\item\label{ax:sodos:clun}
if $A\in K$ and, for all $a\in A$, $t_a\in\closure(\{a\})$, then $\{t_a\colon a\in A\}\in K$
\end{enumerate}
\begin{definition}\label{def:cohsodos2fin}
A set of desirable sets $K\subseteq\setsofthings$ is called \emph{finitely coherent} if it satisfies \ref{ax:sodos:nonempty}--\ref{ax:sodos:des} and~\ref{ax:sodos:clfin}. It is called \emph{$2$-coherent} if it satisfies \ref{ax:sodos:nonempty}--\ref{ax:sodos:des} and~\ref{ax:sodos:clbin}. It is called \emph{$1$-coherent} if it satisfies \ref{ax:sodos:nonempty}--\ref{ax:sodos:des} and~\ref{ax:sodos:clun}.
\end{definition}

A common feature of the previous studies that used a simpler version of \ref{ax:sodos:cl}~\cite{debock2018,pmlr-v103-de-bock19b}, resembling~\ref{ax:sodos:clbin}, is that the desirable sets of gambles they consider are all finite: in those studies, sets of desirable sets (of gambles) are subsets of
\begin{equation*}
\finsetsofthings\coloneqq\{A\in\setsofthings\colon \abs{A}<\infty\}
\end{equation*}
rather than $\setsofthings$. In such a setting, some of the coherence axioms need to be modified slightly, in such a way that they do not enforce us to add infinite sets of things. In particular, $B$ and $\{t_S\colon S\in\mathcal{A}_\mathcal{A}\}$ need to be finite for \ref{ax:sodos:mon} and~\ref{ax:sodos:cl} to make sense, respectively. To enable us to consider this setting as well, we introduce a more general notion of coherence for sets of desirable sets, which explicitly restricts attention to a specific set of sets $\mathcal{Q}\subseteq\setsofthings$, such as for example $\mathcal{Q}=\finsetsofthings$.

\begin{definition}
\label{def:cohsodosQ}
For any $\mathcal{Q}\subseteq\setsofthings$, a set of desirable sets (of things) $K\subseteq\setsofthings$ is \emph{coherent in} $\mathcal{Q}$ if it satisfies
\begin{enumerate}[label=$\mathrm{K}_{\arabic*}^\mathcal{Q}$.,ref=$\mathrm{K}_{\arabic*}^\mathcal{Q}$,leftmargin=*]
\item\label{ax:sodos:nonemptyQ} $\emptyset\notin K$
\item\label{ax:sodos:monQ} if $A\subseteq B\in\mathcal{Q}$ and $A\in K\cap\mathcal{Q}$, then also $B\in K$
\item\label{ax:sodos:notQ} if $A\in K\cap\mathcal{Q}$ and $A\setminus A_\mathrm{not}\in\mathcal{Q}$, then also $A\setminus A_\mathrm{not}\in K$
\item\label{ax:sodos:desQ} $\{t\}\in K$ for all $t\in A_\mathrm{des}$ such that $\{t\}\in\mathcal{Q}$
\item\label{ax:sodos:clQ} if $\emptyset\neq\mathcal{A}\subseteq K\cap\mathcal{Q}$ and, for all $S\in\mathcal{S}_\mathcal{A}$, $t_S\in\closure(S)$, then if $\{t_S\colon S\in\mathcal{S}_\mathcal{A}\}\in\mathcal{Q}$, also $\{t_S\colon S\in\mathcal{S}_\mathcal{A}\}\in K$.
\end{enumerate}
It is called \emph{finitely coherent in} $\mathcal{Q}$ if it satisfies \ref{ax:sodos:nonemptyQ}--\ref{ax:sodos:desQ} and
\begin{enumerate}[label=$\mathrm{K}_{5\mathrm{fin}}^\mathcal{Q}$.,ref=$\mathrm{K}_{5\mathrm{fin}}^\mathcal{Q}$,leftmargin=*]
\item\label{ax:sodos:clfinQ} if $\emptyset\neq\mathcal{A}\subseteq K\cap\mathcal{Q}$ is finite and, for all $S\in\mathcal{S}_\mathcal{A}$, $t_S\in\closure(S)$,\\ then if $\{t_S\colon S\in\mathcal{S}_\mathcal{A}\}\in\mathcal{Q}$, also $\{t_S\colon S\in\mathcal{S}_\mathcal{A}\}\in K$.
\end{enumerate} 
It is called \emph{$2$-coherent in} $\mathcal{Q}$ if it satisfies \ref{ax:sodos:nonemptyQ}--\ref{ax:sodos:desQ} and
\begin{enumerate}[label=$\mathrm{K}_{5\mathrm{bin}}^\mathcal{Q}$.,ref=$\mathrm{K}_{5\mathrm{bin}}^\mathcal{Q}$,leftmargin=*]
\item\label{ax:sodos:clbinQ} if $A,B\in K\cap\mathcal{Q}$ and, for all $a\in A$ and $b\in B$, $t_{a,b}\in\closure(\{a,b\})$, \\then if $\{t_{a,b}\colon a\in A, b\in B\}\in\mathcal{Q}$, also $\{t_{a,b}\colon a\in A, b\in B\}\in K$.
\end{enumerate}
It is called \emph{$1$-coherent in} $\mathcal{Q}$ if it satisfies \ref{ax:sodos:nonemptyQ}--\ref{ax:sodos:desQ} and
\begin{enumerate}[label=$\mathrm{K}_{5\mathrm{un}}^\mathcal{Q}$.,ref=$\mathrm{K}_{5\mathrm{un}}^\mathcal{Q}$,leftmargin=*]
\item\label{ax:sodos:clunQ}
if $A\in K\cap\mathcal{Q}$ and, for all $a\in A$, $t_a\in\closure(\{a\})$, then if $\{t_a\colon a\in A\}\in\mathcal{Q}$,\\ also $\{t_a\colon a\in A\}\in K$
\end{enumerate}
%We denote the set of all coherent sets of desirable things by $\sodoss$.
\end{definition}

If $\mathcal{Q}=\setsofthings$, these new notions of coherence are identical to the ones in Definitions~\ref{def:cohsodos} and~\ref{def:cohsodos2fin}; for example, coherence in $\setsofthings$ is the same as coherence. If $\mathcal{Q}$ is a strict subset of $\setsofthings$, however, the new notions amount to imposing the previous ones only to the extent that they involve sets in $\mathcal{Q}$. For that reason, coherence implies coherence in $\mathcal{Q}$, and similarly for finite coherence (in $\mathcal{Q}$), 2-coherence (in $\mathcal{Q}$) and 1-coherence (in $\mathcal{Q}$). For the important case where $\mathcal{Q}=\finsetsofthings$, some of the axioms simplify slightly because the condition of belonging to $\mathcal{Q}$ becomes redundant. They then reduce to the following versions, where we use a superscript $\mathrm{fin}$ rather then $\mathcal{Q}$ to indicate that we are dealing with the special case $\mathcal{Q}=\finsetsofthings$.
\begin{enumerate}[label=$\mathrm{K}_{\arabic*}^\mathrm{fin}$.,ref=$\mathrm{K}_{\arabic*}^\mathrm{fin}$,leftmargin=*]
\item\label{ax:sodos:nonemptyQfin} $\emptyset\notin K$
\item\label{ax:sodos:monQfin} if $A\subseteq B\in\finsetsofthings$ and $A\in K$, then also $B\in K$
\item\label{ax:sodos:notQfin} if $A\in K\cap\finsetsofthings$ then also $A\setminus A_\mathrm{not}\in K$
\item\label{ax:sodos:desQfin} $\{t\}\in K$ for all $t\in A_\mathrm{des}$
\item\label{ax:sodos:clQfin} if $\emptyset\neq\mathcal{A}\subseteq K\cap\finsetsofthings$ and, for all $S\in\mathcal{S}_\mathcal{A}$, $t_S\in\closure(S)$,\\ then if $\{t_S\colon S\in\mathcal{S}_\mathcal{A}\}\in\finsetsofthings$, also $\{t_S\colon S\in\mathcal{S}_\mathcal{A}\}\in K$,
\end{enumerate}
\begin{enumerate}[label=$\mathrm{K}_{5\mathrm{fin}}^\mathrm{fin}$.,ref=$\mathrm{K}_{5\mathrm{fin}}^\mathrm{fin}$,leftmargin=*]
\item\label{ax:sodos:clfinQfin} if $\emptyset\neq\mathcal{A}\subseteq K\cap\finsetsofthings$ is finite and, for all $S\in\mathcal{S}_\mathcal{A}$, $t_S\in\closure(S)$, then $\{t_S\colon S\in\mathcal{S}_\mathcal{A}\}\in K$.
\end{enumerate} 
\begin{enumerate}[label=$\mathrm{K}_{5\mathrm{bin}}^\mathrm{fin}$.,ref=$\mathrm{K}_{5\mathrm{bin}}^\mathrm{fin}$,leftmargin=*]
\item\label{ax:sodos:clbinQfin} if $A,B\in K\cap\finsetsofthings$ and, for all $a\in A$ and $b\in B$, $t_{a,b}\in\closure(\{a,b\})$, then $\{t_{a,b}\colon a\in A, b\in B\}\in K$.
\end{enumerate}
\begin{enumerate}[label=$\mathrm{K}_{5\mathrm{un}}^\mathrm{fin}$.,ref=$\mathrm{K}_{5\mathrm{un}}^\mathrm{fin}$,leftmargin=*]
\item\label{ax:sodos:clunQfin}
if $A\in K\cap\finsetsofthings$ and, for all $a\in A$, $t_a\in\closure(\{a\})$, then $\{t_a\colon a\in A\}\in K$
\end{enumerate}
For \ref{ax:sodos:clfinQfin}, the reason why there is no need to check whether $\{t_S\colon S\in\mathcal{S}_\mathcal{A}\}\in\finsetsofthings$, is because the finitenes of $\mathcal{A}$ and all $A\in\mathcal{A}$ implies that $\mathcal{S}_\mathcal{A}$ and therefore also $\{t_S\colon S\in\mathcal{S}_\mathcal{A}\}$ is finite. Similarly, for \ref{ax:sodos:clbinQfin}, the finiteness of $A$ and $B$ implies that $\{t_{a,b}\colon a\in A, b\in B\}$ is finite as well, and for \ref{ax:sodos:clunQfin}, the finiteness of $A$ implies that $\{t_a\colon a\in A\}$ is finite.

Regardless of the particular choice of $\mathcal{Q}$, since~\ref{ax:sodos:clQ} trivially implies~\ref{ax:sodos:clfinQ} which in turn implies \ref{ax:sodos:clunQ}, we know that coherence in $\mathcal{Q}$ implies finite coherence in $\mathcal{Q}$ which in turn implies 1-coherence in $\mathcal{Q}$. Similarly, 2-coherence in $\mathcal{Q}$ implies 1-coherence in $\mathcal{Q}$ because \ref{ax:sodos:clbinQ} implies \ref{ax:sodos:clunQ}.\footnote{It suffices to apply \ref{ax:sodos:clbinQ} with $B\coloneqq A$ and, for all $a\in A$ and $b\in B$, $t_{a,b}\coloneqq t_a\in\closure(\{a\})\subseteq\closure(\{a,b\})$, using \ref{ax:clos:mon} for the final inclusion.}
At first sight, one might think that similarly,~\ref{ax:sodos:clfinQ} (and hence also~\ref{ax:sodos:clQ}) trivially implies~\ref{ax:sodos:clbinQ}. This may not be the case though, the subtle issue being that the case $A=B$ may not be implied by~\ref{ax:sodos:clfin}. Nevertheless, finite coherence in $\mathcal{Q}$ does imply 2-coherence in $\mathcal{Q}$, provided that $\mathcal{Q}$ is closed with respect to taking subsets, as is the case for the important cases $\setsofthings$ and $\finsetsofthings$.

\begin{definition}\label{def:closedwrtsubsets}
A set $\mathcal{Q}\subseteq\setsofthings$ is \emph{closed with respect to taking subsets} if for all $A\in\setsofthings$ such that $A\subseteq B\in\mathcal{Q}$, also $A\in\mathcal{Q}$.
\end{definition}

\begin{proposition}\label{prop:fintobinQ}
Let $\mathcal{Q}\subseteq\setsofthings$ be closed with respect to taking subsets.
Let $K\subseteq\setsofthings$ be a set of desirable sets. Then if $K$ satisfies~\ref{ax:sodos:clfinQ} and~\ref{ax:sodos:monQ}, it also satifies~\ref{ax:sodos:clbinQ}. Consequently, if $K$ is finitely coherent in $\mathcal{Q}$, it is also 2-coherent in $\mathcal{Q}$.
\end{proposition}
\begin{proof}
Assume that $K$ satisfies~\ref{ax:sodos:clfinQ} and~\ref{ax:sodos:monQ}. To prove that it then also satifies~\ref{ax:sodos:clbinQ}, we consider any $A,B\in K\cap\mathcal{Q}$ and, for all $a\in A$ and $b\in B$, some $t_{a,b}\in\closure(\{a,b\})$, and assume that $C\coloneqq\{t_{a,b}\colon a\in A, b\in B\}\in\mathcal{Q}$. We need to prove that $C\in K$. If $A\neq B$, this follows directly from \ref{ax:sodos:clfinQ}, with $\mathcal{A}=\{A,B\}$. If $A=B$, then for all $a\in A=B$, we let $t_a\coloneqq t_{a,a}\in\closure(\{a,a\})=\closure(\{a\})$, and let $\tilde{C}\coloneqq\{t_a\colon a\in A\}$. Then since $\tilde{C}\subseteq C\in\mathcal{Q}$ and $\mathcal{Q}$ is closed with respect to taking subsets, we know that $\tilde{C}\in\mathcal{Q}$, and it therefore follows from \ref{ax:sodos:clfinQ}, with $\mathcal{A}=\{A\}$, that $\tilde{C}\in K$. Since $\tilde{C}\subseteq C\in\mathcal{Q}$ and $\tilde{C}\in\mathcal{Q}$, \ref{ax:sodos:monQ} now implies that $C\in K$.
\end{proof}

The more interesting question is whether finite coherence in $\mathcal{Q}$, 2-coherence in $\mathcal{Q}$ or 1-coherence in $\mathcal{Q}$ implies coherence in $\mathcal{Q}$, since that would allow us to replace our axioms with simpler yet equivalent ones. A first important result is that if $\mathcal{Q}$ is closed with respect to taking subsets, then all these notions of coherence are equivalent, provided that the closure operator $\closure$ is unitary. In that case, without loss of power, coherence in $\mathcal{Q}$ can therefore be replaced by 1-coherence in $\mathcal{Q}$.

\begin{definition}
\label{def:unitarycl}
A closure operator $\closure$ is unitary if
\begin{equation*}
\closure(A)=\cup_{t\in A}\closure(\{t\})
\text{ for all }A\in\setsofthings.
\end{equation*}
\end{definition}

\begin{proposition}\label{prop:unitarycl}
Assume that $\closure$ is unitary and that $\mathcal{Q}$ is closed with respect to taking subsets. Then any set of desirable sets $K\subseteq\setsofthings$ that satisfies \ref{ax:sodos:clunQ} and~\ref{ax:sodos:monQ} will also satisfy \ref{ax:sodos:clbinQ}, \ref{ax:sodos:clfinQ} and \ref{ax:sodos:clQ}. Consequently, coherence in $\mathcal{Q}$ is then equivalent to finite coherence in $\mathcal{Q}$, 2-coherence in $\mathcal{Q}$ and 1-coherence in $\mathcal{Q}$.
\end{proposition}
\begin{proof}
Assume that $K$ satisfies~\ref{ax:sodos:clunQ} and~\ref{ax:sodos:monQ}. Since \ref{ax:sodos:clQ} implies \ref{ax:sodos:clfinQ}, and \ref{ax:sodos:clfinQ} and~\ref{ax:sodos:monQ} imply~\ref{ax:sodos:clbinQ} due to Proposition~\ref{prop:fintobinQ}, it suffices to prove \ref{ax:sodos:clQ}.

To prove \ref{ax:sodos:clQ}, consider any $\emptyset\neq\mathcal{A}\subseteq K\cap\mathcal{Q}$ and, for all $S\in\mathcal{S}_\mathcal{A}$, some $t_S\in\closure(S)$, and assume that $B\coloneqq\{t_S\colon S\in\mathcal{S}_\mathcal{A}\}\in\mathcal{Q}$. We need to prove that $B\in K$. For all $S\in\mathcal{S}_\mathcal{A}$, since $\closure$ is unitary, it follows from $t_S\in\closure(S)$ that there is some $\tilde{t}_S\in S$ such that $t_S\in\closure(\{\tilde{t}_S\})$. Let $\tilde{B}\coloneqq\{\tilde{t}_S\colon S\in\mathcal{S}_\mathcal{A}\}$. It then follows from Lemma~\ref{lem:trickforidentityoperator} that there is some $A\in\mathcal{A}$ such that $A\subseteq\tilde{B}$. Consider any such $A$. Then for all $a\in A$, since $a\in A\subseteq\tilde{B}$, there is some $S\in\mathcal{S}_\mathcal{A}$ such that $a=\tilde{t}_S$. Let $t_a\coloneqq t_S$. Then $t_a=t_S\in B$ and, since $t_S\in\closure(\{\tilde{t}_S\})$ and $\tilde{t}_S=a$, also $t_a\in\closure(\{a\})$. 
Doing so for every $a\in A$, we obtain a set $\tilde{A}\coloneqq\{t_a\colon a\in A\}\subseteq B$, which belongs to $\mathcal{Q}$ because $B\in\mathcal{Q}$ and $\mathcal{Q}$ is closed with respect to taking subsets.
It now follows from~\ref{ax:sodos:clunQ} that $\tilde{A}\in K$ and therefore, due to~\ref{ax:sodos:monQ}, that $B\in K$.
\end{proof}

A particular class of unitary closure operators to which this result can be applied, are those that close a set with respect to an equivalence relation. Another are closure operators for which $\closure(A)$ is the upset of $A$ according to some partial order on $\things$. Unfortunately though, being unitary is a very strong property to ask of a closure operator, and only the most simple closure operators will satisfy it. 
The $\posi$ operator that is typically used in the context of desirable gambles---see Example~\ref{ex:gambles}---is for example not unitary, nor is the convex hull operator that appeared in Examples~\ref{ex:nonlineargambles} and~\ref{ex:lotteriesandsuch} or the transitive closure in Example~\ref{example:preferencesasthings}.

Fortunately, we can still obtain equivalences between coherence in $\mathcal{Q}$, finite coherence in $\mathcal{Q}$ and 2-coherence in $\mathcal{Q}$---but not 1-coherence in $\mathcal{Q}$---for closure operators that are not unitary, provided we make some other assumptions. We will establish such equivalences for the particular cases $\mathcal{Q}=\finsetsofthings$ and $\mathcal{Q}=\setsofthings$, starting with the former.

Our first result for the case $\mathcal{Q}=\finsetsofthings$, is that finite coherence in $\finsetsofthings$ is equivalent to coherence in $\finsetsofthings$ provided that the closure operator $\closure$ is finitary. As can be seen from the following definition, these finitary closure operators are a superset of the unitary ones.

\begin{definition}[Finitary closure operator]
\label{def:finclosop}
A closure operator $\closure$ is finitary if
\begin{equation}
\label{ax:clos:fin} 
\closure(A)=\bigcup_{B\subseteq A, \abs{B}<\infty}\closure(B)
\end{equation}
\end{definition}

\begin{corollary}\label{cor:fincohiffcoh:finitecase}
If $\closure$ is finitary, a set of desirable sets $K\subseteq\setsofthings$ is finitely coherent in $\finsetsofthings$ if and only if it is coherent in $\finsetsofthings$.
\end{corollary}
Our proof makes use of the following important technical lemma, which will come in handy later as well in our proof for Proposition~\ref{prop:fromKtofinK}.

\begin{lemma}\label{lem:tychonof}
If $\closure$ is finitary, $\mathcal{A}\subseteq\finsetsofthings$ is non-empty and, for all $S\in\mathcal{S}_\mathcal{A}$, $t_S\in\closure(S)$, then there is some finite non-empty $\mathcal{A}_\mathrm{fin}\subseteq\mathcal{A}$ and, for all $S\in\mathcal{S}_{\mathcal{A}_\mathrm{fin}}$, some $t^\mathrm{fin}_S\in\closure(S)$ such that $\{t^\mathrm{fin}_S\colon S\in\mathcal{S}_{\mathcal{A}_\mathrm{fin}}\}$ is a finite subset of $\{t_S\colon S\in\mathcal{S}_\mathcal{A}\}$.
\end{lemma}
\begin{proof}
Assume that $\closure$ is finitary, and consider any non-empty $\mathcal{A}\subseteq\finsetsofthings$ and, for all $S\in\mathcal{S}_\mathcal{A}$, some $t_S\in\closure(S)$. Let $C\coloneqq\{t_S\colon S\in\mathcal{S}_\mathcal{A}\}$. Our proof distinguishes two cases: $\emptyset\in\mathcal{A}$ and $\emptyset\notin\mathcal{A}$. 

If $\emptyset\in\mathcal{A}$, then $\mathcal{S}_\mathcal{A}=\emptyset$ and therefore also $\{t_S\colon S\in\mathcal{S}_\mathcal{A}\}=\emptyset$. In that case, we let $\mathcal{A}_\mathrm{fin}\coloneqq\{\emptyset\}$, which is then a finite non-empty subset of $\mathcal{A}$. Furthermore, we also have that $\mathcal{S}_{\mathcal{A}_\mathrm{fin}}=\emptyset$, and therefore, that $\{t^\mathrm{fin}_S\colon S\in\mathcal{S}_{\mathcal{A}_\mathrm{fin}}\}=\emptyset=\{t_S\colon S\in\mathcal{S}_\mathcal{A}\}$.

So it remains to consider the case $\emptyset\notin\mathcal{A}$. In that case, let $\Phi_\mathcal{A}$ be the cartesian product of the set $\mathcal{A}$: the set of all maps $\phi\colon\mathcal{A}\to\things$ such that, for all $A\in\mathcal{A}$, $\phi(A)\in A$.
We endow every $A\in\mathcal{A}$ with the discrete topology and endow $\Phi_\mathcal{A}$ with the product topology. Since $\mathcal{A}\subseteq\finsetsofthings$ and $\emptyset\notin\mathcal{A}$, every $A\in\mathcal{A}$ is finite and non-empty and therefore trivially compact with respect to its discrete topology. It therefore follows from Tychonoff's theorem that $\Phi_\mathcal{A}$ is compact with respect to its product topology. Furthermore, since $\emptyset\notin\mathcal{A}$, we know that $\Phi_\mathcal{A}\neq\emptyset$ by the axiom of choice.

Now observe that $\mathcal{S}_\mathcal{A}=\{\{\phi(A)\colon A\in\mathcal{A}\}\colon \phi\in\Phi_\mathcal{A}\}$. Therefore, for all $\phi\in\Phi_\mathcal{A}$, there is some $S_\phi\in\mathcal{S}_\mathcal{A}$ such that $S_\phi=\{\phi(A)\colon A\in\mathcal{A}\}$. Since $t_{S_\phi}\in\closure(S_\phi)$ and $\closure$ is finitary, there is some finite $B_\phi\subseteq S$ such that $t_{S_\phi}\in\closure(B_\phi)$. Due to~\ref{ax:clos:empty}, $B_\phi$ is then necessarily non-empty. 
Since $B_\phi\subseteq S_\phi=\{\phi(A)\colon A\in\mathcal{A}\}$ is finite and non-empty, there is a finite non-empty set $\mathcal{A}_\phi\subseteq\mathcal{A}$ such that $B_\phi=\{\phi(A)\colon A\in\mathcal{A}_\phi\}$. 
Let $\Phi_\phi\coloneqq\{\phi^*\in\Phi_\mathcal{A}\colon\phi^*(A)=\phi(A)\text{ for all }A\in\mathcal{A}_\phi\}$. Then $\phi\in\Phi_\phi$ and, since $\mathcal{A}_\phi$ is finite, $\Phi_\phi$ is open (since it is a cylinder set) in the product topology.

Since $\phi\in\Phi_\phi$ for all $\phi\in\Phi_\mathcal{A}$, we know that $\Phi_\mathcal{A}=\cup_{\phi\in\Phi_\mathcal{A}}\{\phi\}\subseteq\cup_{\phi\in\Phi_\mathcal{A}}\Phi_\phi$, so $\{\Phi_\phi\colon \phi\in\Phi_\mathcal{A}\}$ is an open cover of $\Phi_\mathcal{A}$. Since $\Phi_\mathcal{A}$ is compact, this implies it has a finite subcover. That is, there is some finite set $\Phi_\mathrm{fin}\subseteq\Phi_\mathcal{A}$ such that $\{\Phi_\phi\colon \phi\in\Phi_\mathrm{fin}\}$ is a cover of $\Phi_\mathcal{A}$, meaning that $\Phi_\mathcal{A}\subseteq\cup_{\phi\in\Phi_\mathrm{fin}}\Phi_\phi$. Since $\Phi_\mathcal{A}\neq\emptyset$, this also implies that $\Phi_\mathrm{fin}$ is non-empty. Let $\mathcal{A}_\mathrm{fin}\coloneqq\cup_{\phi\in\Phi_\mathrm{fin}}\mathcal{A}_\phi\subseteq\mathcal{A}$. Then $\mathcal{A}_\mathrm{fin}$ is finite and non-empty because $\Phi_\mathrm{fin}$ and all $\mathcal{A}_\phi$ are finite and non-empty. This also implies that $\mathcal{S}_{\mathcal{A}_\mathrm{fin}}$ is finite because all $A\in\mathcal{A}_\mathrm{fin}\subseteq\mathcal{A}\subseteq\finsetsofthings$ are finite.

For all $S\in\mathcal{S}_{\mathcal{A}_\mathrm{fin}}$, we know that $S=\{t^*_A\colon A\in\mathcal{A}_\mathrm{fin}\}$, with $t^*_{A}\in A$ for all $A\in\mathcal{A}_\mathrm{fin}$. Consider any $\phi^*\in\Phi_\mathcal{A}$ such that $\phi^*(A)=t^*_A$ for all $A\in\mathcal{A}_\mathrm{fin}\subseteq\mathcal{A}$ (since $\emptyset\notin\mathcal{A}$, such a $\phi^*$ exists by the axiom of choice). Since $\phi^*\in\Phi_\mathcal{A}\subseteq\cup_{\phi\in\Phi_\mathrm{fin}}\Phi_\phi$, there is some $\phi\in\Phi_\mathrm{fin}$ such that $\phi^*\in\Phi_\phi$. Let $t^\mathrm{fin}_S\coloneqq t_{S_\phi}$. Since $\phi^*\in\Phi_\phi$, we know that $\phi^*(A)=\phi(A)$ for all $A\in\mathcal{A}_\phi$, which implies that
\begin{align*}
B_\phi
=\{\phi(A)\colon A\in\mathcal{A}_\phi\}
&=\{\phi^*(A)\colon A\in\mathcal{A}_\phi\}\\
&\subseteq\{\phi^*(A)\colon A\in\mathcal{A}_\mathrm{fin}\}
=\{t^*_A\colon A\in\mathcal{A}_\mathrm{fin}\}=S.
\end{align*}
Since $t^\mathrm{fin}_S=t_{S_\phi}\in\closure(B_\phi)$, it therefore follows from~\ref{ax:clos:mon} that $t^\mathrm{fin}_S\in\closure(S)$. On the other hand, since $S_\phi\in\mathcal{S}_\mathcal{A}$, we also know that $t^\mathrm{fin}_S=t_{S_\phi}\in C$. Hence, for all $S\in\mathcal{S}_{\mathcal{A}_\mathrm{fin}}$, we have found some $t^\mathrm{fin}_S\in\closure(S)$ such that $t^\mathrm{fin}_S\in C$. This implies that $\tilde{C}\coloneqq\{t^\mathrm{fin}_S\colon S\in\mathcal{S}_{\mathcal{A}_\mathrm{fin}}\}\subseteq C$, where $\tilde{C}$ is finite because $\mathcal{S}_{\mathcal{A}_\mathrm{fin}}$ is.
\end{proof}

\begin{proof}[Proof of Corollary~\ref{cor:fincohiffcoh:finitecase}]
Since coherence in $\finsetsofthings$ trivially implies finite coherence in $\finsetsofthings$, it suffices to prove the converse. So assume that $K$ is finitely coherent in $\finsetsofthings$, meaning that it satisfies~\ref{ax:sodos:nonemptyQfin}--\ref{ax:sodos:desQfin} and \ref{ax:sodos:clfinQfin}, and that $\closure$ is finitary. We need to prove that $K$ then also satisfies \ref{ax:sodos:clQfin}. To that end, consider any $\emptyset\neq\mathcal{A}\subseteq K\cap\finsetsofthings$ and, for all $S\in\mathcal{S}_\mathcal{A}$, some $t_S\in\closure(S)$, and assume that $C\coloneqq\{t_S\colon S\in\mathcal{S}_\mathcal{A}\}\in\finsetsofthings$. We need to prove that $C\in K$.
Due to Lemma~\ref{lem:tychonof}, there is some finite non-empty $\mathcal{A}_\mathrm{fin}\subseteq\mathcal{A}$ and, for all $S\in\mathcal{S}_{\mathcal{A}_\mathrm{fin}}$, some $t^\mathrm{fin}_S\in\closure(S)$ such that $\tilde{C}\coloneqq\{t^\mathrm{fin}_S\colon S\in\mathcal{S}_{\mathcal{A}_\mathrm{fin}}\}$ is a subset of $C$. 
Since $\emptyset\neq\mathcal{A}_\mathrm{fin}\subseteq\mathcal{A}\subseteq K\cap\finsetsofthings$ and $\mathcal{A}_\mathrm{fin}$ is finite, it follows from~\ref{ax:sodos:clfinQfin} that $\tilde{C}\in K$. Since $\tilde{C}$ is a subset of $C\in\finsetsofthings$, it therefore follows from~\ref{ax:sodos:monQfin} that $C\in K$.
\end{proof}

Similarly, it is also possible to replace finite coherence in $\finsetsofthings$ with 2-coherence in $\finsetsofthings$. For that result, the closure operator $\closure$ needs to be incremental. Here too, it can easily be seen that unitary closure operators are always incremental.

\begin{definition}[Incremental closure operator]
\label{def:incrclosop}
A closure operator is \emph{incremental} if for all $A\in\setsofthings$, $a\in\things$ and $t\in\closure(A\cup\{a\})$, there is some $t_A\in\closure(A)$ such that $t\in\closure(\{t_A,a\})$. 
\end{definition}

\begin{proposition}\label{fincohifbincoh:finitecase}
If $\closure$ is incremental, then any set of desirable sets $K\subseteq\setsofthings$ that satisfies \ref{ax:sodos:clbinQfin} and \ref{ax:sodos:monQfin} will also satisfy \ref{ax:sodos:clfinQfin}. Consequently, finite coherence in $\finsetsofthings$ is then equivalent to 2-coherence in $\finsetsofthings$.
\end{proposition}
\begin{proof}
Since $\finsetsofthings$ is closed with respect to taking subsets, the second part of the result follows from the first part and Proposition~\ref{prop:fintobinQ}. We therefore only prove the first part. To that end, assume that $\closure$ is incremental and consider any set of desirable sets $K$ that satisfies~\ref{ax:sodos:clbinQfin} and~\ref{ax:sodos:monQfin}. To prove that $K$ satisfies~\ref{ax:sodos:clfinQfin}, it clearly suffices to prove that for all $n\in\mathbb{N}$ (the natural numbers excluding zero), $K$ satisfies
\begin{enumerate}[label=$\mathrm{K}_{5,n}^\mathrm{fin}$.,ref=$\mathrm{K}_{5,n}^\mathrm{fin}$,leftmargin=*]
\item if $\mathcal{A}\subseteq K\cap\finsetsofthings$ with $\abs{\mathcal{A}}=n$ and, for all $S\in\mathcal{S}_\mathcal{A}$, $t_S\in\closure(S)$, \\then $\{t_S\colon S\in\mathcal{S}_\mathcal{A}\}\in K$.
\end{enumerate}
We first prove that $K$ satisfies $\mathrm{K}_{5,n}^\mathrm{fin}$ for $n=1$. To that end, consider $\mathcal{A}\subseteq K\cap\finsetsofthings$ with $\abs{\mathcal{A}}=1$ and, for all $S\in\mathcal{S}_\mathcal{A}$, some $t_S\in\closure(S)$. 
We need to prove that then $C\coloneqq\{t_S\colon S\in\mathcal{S}_\mathcal{A}\}\in K$. Since $\abs{\mathcal{A}}=1$ and $\mathcal{A}\subseteq K\cap\finsetsofthings$, we know that $\mathcal{A}=\{A\}$ for some finite $A\in K$. Observe that $\mathcal{S}_\mathcal{A}=\{\{a\}\colon a\in A\}$, which is finite because $A$ is, and that $C$ is therefore finite as well. Let $B\coloneqq A$. For all $a\in A$ and $b\in B$, let $t_{a,b}\coloneqq t_{\{a\}}\in C$. It then follows from~\ref{ax:clos:mon} that $t_{a,b}=t_{\{a\}}\in\closure(\{a\})\subseteq\closure(\{a,b\})$. Since $B=A\in K$ is finite, it therefore follows from~\ref{ax:sodos:clbinQfin} that $\{t_{a,b}\colon a\in A,b\in B\}\in K$. Since $\{t_{a,b}\colon a\in A,b\in B\}\subseteq C$ and $C$ is finite, \ref{ax:sodos:monQfin} now implies that $C\in K$.

To show that $K$ satisfies $\mathrm{K}_{5,n}^\mathrm{fin}$ also for $n>1$, we provide a proof by induction. That is, we assume that $K$ satisfies $\mathrm{K}_{4,n}^\mathrm{fin}$ for some $n\in\mathbb{N}$ and prove that it satisfies $\mathrm{K}_{5,n+1}^\mathrm{fin}$.
To prove $\mathrm{K}_{5,n+1}^\mathrm{fin}$, we consider any $\mathcal{A}_{n+1}\subseteq K\cap\finsetsofthings$ with $\abs{\mathcal{A}_{n+1}}=n+1$ and, for all $S\in\mathcal{S}_{\mathcal{A}_{n+1}}$, some $t_S\in\closure(S)$. 
We again need to show that $C\coloneqq\{t_S\colon S\in\mathcal{S}_{\mathcal{A}_{n+1}}\}\in K$. 
Note that, since $\mathcal{A}_{n+1}$ and all $A\in\mathcal{A}_{n+1}$ are finite, $\mathcal{S}_{\mathcal{A}_{n+1}}$ and therefore also $C$ are finite as well.
Consider any $A_{n+1}\in\mathcal{A}_{n+1}$ and let $\mathcal{A}_n\coloneqq\mathcal{A}_{n+1}\setminus\{A_{n+1}\}$. Then $\abs{\mathcal{A}_n}=n$. 
Furthermore, since $A_{n+1}\in\mathcal{A}_{n+1}\subseteq K$ and $C$ are finite, it follows from~\ref{ax:sodos:monQfin} that also $A_{n+1}\cup C\in K$. To prove that $C\in K$, we will now show that for any $B\subseteq A_{n+1}$ such that $B\cup C\in K$ and any $b\in B$, also $(B\setminus\{b\})\cup C\in K$. Since $A_{n+1}\in\finsetsofthings$ is finite and $A_{n+1}\cup C\in K$, repeated application of this result then indeed implies that $C\in K$.

So consider any $B\subseteq A_{n+1}$ such that $B\cup C\in K$ and any $b\in B$. For all $S\in\mathcal{S}_{\mathcal{A}_n}$, since $b\in B\subseteq A_{n+1}$ and $\mathcal{A}_n=\mathcal{A}_{n+1}\setminus\{A_{n+1}\}$, we know that $S\cup\{b\}\in\mathcal{S}_{\mathcal{A}_{n+1}}$, hence also $t_{S\cup\{b\}}\in\closure(S\cup\{b\})$. Since $\closure$ is incremental, this implies that there is some $t^*_S\in\closure(S)$ such that $t_{S\cup\{b\}}\in\closure(\{t^*_S,b\})$. Let $C^*\coloneqq\{t^*_S\colon S\in\mathcal{S}_{\mathcal{A}_n}\}$. Then since $\mathcal{A}_{n}$ and all $A\in\mathcal{A}_{n}$ are finite, $\mathcal{S}_{\mathcal{A}_{n+1}}$ and therefore also $C^*$ are finite. Furthermore, since $\mathcal{A}_n\subseteq\mathcal{A}_{n+1}\subseteq K\cap\finsetsofthings$ and $\abs{\mathcal{A}_n}=n$, it follows from the induction hypothesis (that $K$ satisfies $\mathrm{K}_{4,n}^\mathrm{fin}$)  that $C^*\in K$.
For any $c^*\in C^*$ and $t\in B\cup C$, we will now define some $t_{c^*,t}\in\things$. We consider two cases: $t\in (B\setminus\{b\})\cup C$ and $t\notin (B\setminus\{b\})\cup C$. If $t\in (B\setminus\{b\})\cup C$, we let $t_{c^*,t}\coloneqq t$. Then on the one hand, $t_{c^*,t}=t\in(B\setminus\{b\})\cup C$. On the other hand, due to \ref{ax:clos:ext}, $t_{c^*,t}=t\in\{c^*,t\}\subseteq\closure(\{c^*,t\})$. If $t\notin (B\setminus\{b\})\cup C$, then $t=b$ because $t\in B\cup C$. Since $c^*\in C^*$, there is some $S\in\mathcal{S}_{\mathcal{A}_n}$ such that $c^*=t^*_S$. We now let $t_{c^*,t}\coloneqq t_{S\cup\{b\}}$. 
Then on the one hand, $t_{c^*,t}=t_{S\cup\{b\}}\in C\subseteq(B\setminus\{b\})\cup C$. 
On the other hand, $t_{c^*,t}\coloneqq t_{S\cup\{b\}}\in\closure(\{t^*_S,b\})=\closure(\{c^*,t\})$. 
Hence, for all $c^*\in C^*$ and $t\in B\cup C$, we have found some $t_{c^*,t}\in(B\setminus\{b\})\cup C$ such that $t_{c^*,t}\in\closure(\{c^*,t\})$. 
Since $C^*\in K$ and $B\cup C\in K$ are both finite, it therefore follows from \ref{ax:sodos:clbinQfin} that $\{t_{c^*,t}\colon c^*\in C^*,t\in B\cup C\}\in K$. Hence, since $\{t_{c^*,t}\colon c^*\in C^*,t\in B\cup C\}\subseteq (B\setminus\{b\})\cup C$ and $(B\setminus\{b\})\cup C\subseteq B\cup C$ is finite, it follows from~\ref{ax:sodos:monQfin} that $(B\setminus\{b\})\cup C\in K$.
\end{proof}

Combining the preceding two results, we arrive at a sufficient condition for coherence in $\finsetsofthings$ to be equivalent to finite coherence in $\finsetsofthings$ and 2-coherence in $\finsetsofthings$.

\begin{corollary}\label{corol:equivalencefincoherence:finitaryincremental}
If $\closure$ is finitary and incremental, then coherence in $\finsetsofthings$ is equivalent to 2-coherence in $\finsetsofthings$, and to finite coherence in $\finsetsofthings$.
\end{corollary}
\begin{proof}
Immediate consequence of Corollary~\ref{cor:fincohiffcoh:finitecase} and Proposition~\ref{fincohifbincoh:finitecase}.
\end{proof}

Unlike the condition that $\closure$ should be unitary, requiring that it is finitary and incremental is not that much to ask. For example, it is easy to verify that the $\posi$ operator in Example~\ref{ex:gambles}, the convex hull operator in Examples~\ref{ex:nonlineargambles} and~\ref{ex:lotteriesandsuch}, and the transitive closure in Example~\ref{example:preferencesasthings}, are all examples of closure operators that are both finitary and incremental.

We have at this point established several conditions under which coherence in $\finsetsofthings$ can be replaced by a simpler alternative. 
The first of these results---Proposition~\ref{prop:unitarycl}---also applies to the case $\mathcal{Q}=\setsofthings$ that we have been considering in the earlier sections of this paper, but the others do not, which is unfortunate because that is the case to which Theorem~\ref{theo:infinite:coherentrepresentation:twosided} applies. To obtain similar results also for the case $\mathcal{Q}=\setsofthings$, we will impose an additional condition, this time on $K$ instead of $\closure$: we will focus on finitary $K$.

\begin{definition}
\label{def:finitarysodos}
A set of desirable sets $K\subseteq\setsofthings$ is called \emph{finitary} if, for all $A\subseteq\mathcal{T}$, we have that $A\in K$ if and only if there is a finite $B\subseteq A$ such that $B\in K$.
\end{definition}

Any such finitary set of desirable sets is completely determined by its restriction to $\finsetsofthings$. Furthermore, as we will see further on, the coherence of a finitary set of desirable sets is also determined by its restriction to $\finsetsofthings$, at least if we impose some suitable conditions.

To show this, we start by defining, for any set of desirable sets $K\subseteq\setsofthings$, the set of all supersets of its restriction to $\finsetsofthings$:
\begin{equation*}
\fin(K)\coloneqq\{A\in\setsofthings\colon B\in K\cap\finsetsofthings, B\subseteq A\}.
\end{equation*}
It is easy to see that $K$ is finitary if and only if it coincides with $\fin(K)$. In fact, even for $K$ that are not finitary, $\fin(K)$ is always finitary.

\begin{proposition}\label{prop:finKalwaysfinitary}
For any set of desirable sets $K\subseteq\setsofthings$, $\fin(K)$ is a finitary set of desirable sets.
\end{proposition}
\begin{proof}
First consider any $A\subseteq\things$ such that $A\in\fin(K)$. Then there is some finite $B\in K$ such that $B\subseteq A$. Since $B\in K$ is finite and $B\subseteq B$, we then also have that $B\in\fin(K)$.
Conversely, consider any $A\subseteq\things$ and any finite $B\subseteq A$ such that $B\in\fin(K)$. Since $B\in\fin(K)$, there is some finite $\tilde{B}\in K$ such that $\tilde{B}\subseteq B$. Since $\tilde{B}\subseteq B\subseteq A$, this implies that $A\in\fin(K)$.
It therefore follows from Definition~\ref{def:finitarysodos} that $\fin(K)$ is finitary.
\end{proof}

\begin{proposition}\label{prop:finitarywithfin}
A set of desirable sets $K\subseteq\setsofthings$ is finitary if and only if $K=\fin(K)$.
\end{proposition}

\begin{proof}
First assume that $K$ is finitary. Then for any $A\in K$, we know from Definition~\ref{def:finitarysodos} that there is some $B\in K\cap\finsetsofthings$ such that $B\subseteq A$, which implies that $A\in\fin(K)$. Conversely, for any $A\in\fin(K)$, there is some $B\in K\cap\finsetsofthings$ such that $B\subseteq A$, so Definition~\ref{def:finitarysodos} implies that $A\in K$. Hence, $K=\fin(K)$.
Next, assume that $K=\fin(K)$. Since $\fin(K)$ is finitary because of Proposition~\ref{prop:finKalwaysfinitary}, $K$ is then finitary as well.
\end{proof}

Due to this result, establishing that a finitary set of desirable sets $K$ satisfies---some version of---coherence is equivalent to establishing that $\fin(K)$ does. Our next result provides some ways for doing just that.

\begin{proposition}\label{prop:fromKtofinK}
For any set of desirable sets $K\subseteq\setsofthings$, we have that
\begin{enumerate}[label=(\roman*)]
\item\label{prop:fromKtofinK:nonempty}
if $K$ satisfies~\ref{ax:sodos:nonemptyQfin}, then $\fin(K)$ satisfies~\ref{ax:sodos:nonempty}
\item\label{prop:fromKtofinK:mon}
$\fin(K)$ satisfies~\ref{ax:sodos:mon}
\item\label{prop:fromKtofinK:finKrestrequalsKrestr}
if $K$ satisfies~\ref{ax:sodos:monQfin}, then $\fin(K)\cap\finsetsofthings=K\cap\finsetsofthings$
\item\label{prop:fromKtofinK:not}
if $K$ satisfies~\ref{ax:sodos:notQfin}, then $\fin(K)$ satisfies~\ref{ax:sodos:not}
\item\label{prop:fromKtofinK:des}
if $K$ satisfies~\ref{ax:sodos:desQfin}, then $\fin(K)$ satisfies~\ref{ax:sodos:des}
\item\label{prop:fromKtofinK:clun}
if $K$ satisfies~\ref{ax:sodos:clunQfin}, then $\fin(K)$ satisfies~\ref{ax:sodos:clun}
\item\label{prop:fromKtofinK:clbin}
if $K$ satisfies~\ref{ax:sodos:clbinQfin}, then $\fin(K)$ satisfies~\ref{ax:sodos:clbin}
\item\label{prop:fromKtofinK:clfin}
if $K$ satisfies~\ref{ax:sodos:clfinQfin}, then $\fin(K)$ satisfies~\ref{ax:sodos:clfin}
\item\label{prop:fromKtofinK:cl}
if $K$ satisfies~\ref{ax:sodos:clfinQfin} and $\closure$ is finitary, then $\fin(K)$ satisfies \ref{ax:sodos:cl}
\end{enumerate}
Consequently, if $K$ is finitely coherent in $\finsetsofthings$, then $\fin(K)$ is finitely coherent and $\fin(K)\cap\finsetsofthings=K\cap\finsetsofthings$, and similarly for 2-coherence and 1-coherence. Furthermore, if $K$ is finitely coherent in $\finsetsofthings$ and $\closure$ is finitary, then $\fin(K)$ is coherent and $\fin(K)\cap\finsetsofthings=K\cap\finsetsofthings$.
\end{proposition}
\begin{lemma}\label{lem:AtoBforfinitaryK}
For any $\emptyset\neq\mathcal{A}\subseteq\fin(K)$, there is some $\emptyset\neq\mathcal{B}\subseteq K\cap\finsetsofthings$ such that $\mathcal{S}_\mathcal{B}\subseteq\mathcal{S}_\mathcal{A}$. Furthermore, if $\mathcal{A}$ is finite, we can guarantee that $\mathcal{B}$ is finite as well.
\end{lemma}
\begin{proof}
For all $A\in\mathcal{A}$, since $\mathcal{A}\subseteq\fin(K)$, there is some $B_A\in\finsetsofthings$ such that $B_A\in K$ and $B_A\subseteq A$. Let $\mathcal{B}\coloneqq\{B_A\colon A\in\mathcal{A}\}\subseteq K\cap\finsetsofthings$, which is non-empty because $\mathcal{A}$ is non-empty.
Then for all $S\in\mathcal{S}_\mathcal{B}$, we know that $S=\{t^*_B\colon B\in\mathcal{B}\}$, with $t^*_B\in B$ for all $B\in\mathcal{B}$. 
Since $\mathcal{B}=\{B_A\colon A\in\mathcal{A}\}$, this implies that $S=\{t^*_B\colon B\in\mathcal{B}\}=\{t^*_{B_A}\colon A\in\mathcal{A}\}$, with $t^*_{B_A}\in B_A\subseteq A$ for all $A\in\mathcal{A}$, and therefore that $S\in\mathcal{S}_\mathcal{A}$. 
So $\mathcal{S}_\mathcal{B}\subseteq\mathcal{S}_\mathcal{A}$. Furthermore, if $\mathcal{A}$ is finite, then $\mathcal{B}=\{B_A\colon A\in\mathcal{A}\}$ is clearly finite as well.
\end{proof}
\begin{proof}[Proof of Proposition~\ref{prop:fromKtofinK}]
For \ref{prop:fromKtofinK:nonempty}, assume that $K$ satisfies~\ref{ax:sodos:nonemptyQfin}. To prove that $\fin(K)$ satisfies \ref{ax:sodos:nonempty}, assume \emph{ex absurdo} that $\emptyset\in\fin(K)$. That implies that there is some $B\in K\cap\finsetsofthings$ such that $B\subseteq\emptyset$. Since $B\subseteq\emptyset$ implies that $B=\emptyset$, it follows that $\emptyset\in K$, contradicting~\ref{ax:sodos:nonemptyQfin}.

For \ref{prop:fromKtofinK:mon}, to prove that $\fin(K)$ satisfies \ref{ax:sodos:mon}, let $A\subseteq B$ and $A\in\fin(K)$. We need to show that $B\in\fin(K)$. Since $A\in\fin(K)$, there is some finite $A^*\in K$ such that $A^*\subseteq A$. Since $A^*\in K$ is finite and $A^*\subseteq A\subseteq B$ this implies that $B\in K$.

For \ref{prop:fromKtofinK:finKrestrequalsKrestr}, assume that $K$ satisfies~\ref{ax:sodos:monQfin}. For all $A\in K\cap\finsetsofthings$, we trivially have that $A\in\fin(K)$, and therefore that $A\in\fin(K)\cap\finsetsofthings$. Conversely, consider any $A\in\fin(K)\cap\finsetsofthings$. Since $A\in\fin(K)$, there is some finite $B\in K$ such that $B\subseteq A$. Since $A\in\finsetsofthings$, it therefore follows from \ref{ax:sodos:monQfin} that $A\in K$, and therefore that $A\in K\cap\finsetsofthings$.

For \ref{prop:fromKtofinK:not}, assume that $K$ satisfies~\ref{ax:sodos:notQfin}. To prove that $\fin(K)$ satisfies \ref{ax:sodos:not}, consider any $A\in\fin(K)$. We need to prove that then also $A\setminus A_\mathrm{not}\in\fin(K)$. Since $A\in\fin(K)$, there is some finite $B\in K$ such that $B\subseteq A$. Since $B\in K$ is finite, it follows from~\ref{ax:sodos:notQfin} that also $B\setminus A_\mathrm{not}\in K$. Observe that $B\setminus A_\mathrm{not}$ is finite because $B$ is, and that $B\setminus A_\mathrm{not}\subseteq A\setminus A_\mathrm{not}$ because $B\subseteq A$. Since $B\setminus A_\mathrm{not}\in K$, it therefore follows that $A\setminus A_\mathrm{not}\in\fin(K)$.

For \ref{prop:fromKtofinK:des}, assume that $K$ satisfies~\ref{ax:sodos:desQfin}. To prove that $\fin(K)$ satisfies \ref{ax:sodos:des}, consider any $t\in A_\mathrm{des}$. We need to show that $\{t\}\in\fin(K)$. Since $K$ satisfies~\ref{ax:sodos:desQfin}, we know that $\{t\}\in K$. Since $\{t\}$ is finite, it follows that also $\{t\}\in\fin(K)$.

For~\ref{prop:fromKtofinK:clun}, assume that $K$ satisfies \ref{ax:sodos:clunQfin}. To prove that $\fin(K)$ satisfies~\ref{ax:sodos:clun}, consider any $A\in\fin(K)$ and, for all $a\in A$, some $t_a\in\closure(\{a\})$. We need to show that $C\coloneqq\{t_a\colon a\in A\}\in\fin(K)$. Since $A\in\fin(K)$, there is some finite $B\in K$ such that $B\subseteq A$. Let $\tilde{C}\coloneqq\{t_a\colon a\in B\}\subseteq C$. Since $B\in K$ is finite, it follows from \ref{ax:sodos:clunQfin} that $\tilde{C}\in K$. Since $\tilde{C}$ is finite because $B$ is, and $\tilde{C}\subseteq C$, this implies that $C\in\fin(K)$.

For~\ref{prop:fromKtofinK:clbin}, assume that $K$ satisfies \ref{ax:sodos:clbinQfin}. To prove that $\fin(K)$ satisfies~\ref{ax:sodos:clbin}, consider any $A,B\in\fin(K)$ and, for all $a\in A$ and $b\in B$, $t_{a,b}\in\closure(\{a,b\})$. We need to show that $C\coloneqq\{t_{a,b}\colon a\in A, b\in B\}\in K$. Since $A,B\in\fin(K)$, there are finite $\tilde{A},\tilde{B}\in K$ such that $\tilde{A}\subseteq A$ and $\tilde{B}\subseteq B$. Let $\tilde{C}\coloneqq\{t_{a,b}\colon a\in\tilde{A}, b\in\tilde{B}\}\subseteq C$. Since $A,B\in K$ are finite, it follows from \ref{ax:sodos:clbinQfin} that $\tilde{C}\in K$. Since $\tilde{C}$ is finite because $\tilde{A}$ and $\tilde{B}$ are, and $\tilde{C}\subseteq C$, this implies that $C\in\fin(K)$.

For~\ref{prop:fromKtofinK:clfin}, assume that $K$ satisfies \ref{ax:sodos:clfinQfin}. To prove that $\fin(K)$ satisfies~\ref{ax:sodos:clfin}, consider any finite $\emptyset\neq\mathcal{A}\subseteq\fin(K)$ and, for all $S\in\mathcal{S}_\mathcal{A}$, some $t_S\in\closure(S)$, and let $C\coloneqq\{t_S\colon S\in\mathcal{S}_\mathcal{A}\}$. We need to show that $C\in K$. Due to Lemma~\ref{lem:AtoBforfinitaryK}, there is some finite $\emptyset\neq\mathcal{B}\subseteq K\cap\finsetsofthings$ such that $\mathcal{S}_\mathcal{B}\subseteq\mathcal{S}_\mathcal{A}$, and therefore $\tilde{C}\coloneqq\{t_S\colon S\in\mathcal{S}_\mathcal{B}\}\subseteq C$. Since $\emptyset\neq\mathcal{B}\subseteq K\cap\finsetsofthings$ is finite and $t_S\in\closure(S)$ for all $S\in\mathcal{S}_\mathcal{B}\subseteq\mathcal{S}_\mathcal{A}$, it now follows from \ref{ax:sodos:clfinQfin} that $\tilde{C}\in K$. Since $\tilde{C}$ is finite because $\mathcal{B}$ and all $B\in\mathcal{B}\subseteq\finsetsofthings$ are finite, this implies that $C\in\fin(K)$ because $\tilde{C}\subseteq C$.

For~\ref{prop:fromKtofinK:cl}, assume that $K$ satisfies \ref{ax:sodos:clfinQfin} and $\closure$ is finitary. To prove that $\fin(K)$ satisfies~\ref{ax:sodos:cl}, consider any $\emptyset\neq\mathcal{A}\subseteq\fin(K)$ and, for all $S\in\mathcal{S}_\mathcal{A}$, some $t_S\in\closure(S)$, and let $C\coloneqq\{t_S\colon S\in\mathcal{S}_\mathcal{A}\}$. We need to show that $C\in K$. 
Due to Lemma~\ref{lem:AtoBforfinitaryK}, there is some $\emptyset\neq\mathcal{B}\subseteq K\cap\finsetsofthings$ such that $\mathcal{S}_\mathcal{B}\subseteq\mathcal{S}_\mathcal{A}$, and therefore $\tilde{C}\coloneqq\{t_S\colon S\in\mathcal{S}_\mathcal{B}\}\subseteq C$. Since $t_S\in\closure(S)$ for all $S\in\mathcal{S}_\mathcal{B}\subseteq\mathcal{S}_\mathcal{A}$, it furthermore follows from Lemma~\ref{lem:tychonof} that there some finite non-empty $\mathcal{B}_\mathrm{fin}\subseteq\mathcal{B}$ and, for all $S\in\mathcal{S}_{\mathcal{B}_\mathrm{fin}}$, some $t^\mathrm{fin}_S\in\closure(S)$, such that $\tilde{C}^*\coloneqq\{t^\mathrm{fin}_S\colon S\in\mathcal{S}_{\mathcal{B}_\mathrm{fin}}\}$ is a finite subset of $\tilde{C}$.
Since $\emptyset\neq\mathcal{B}_\mathrm{fin}\subseteq\mathcal{B}\subseteq K\cap\finsetsofthings$ is finite and $t^\mathrm{fin}_S\in\closure(S)$ for all $S\in\mathcal{S}_{\mathcal{B}_\mathrm{fin}}$, it now follows from \ref{ax:sodos:clfinQfin} that $\tilde{C}^*\in K$. Since $\tilde{C}^*$ is finite because $\mathcal{B}_\mathrm{fin}$ and all $B\in\mathcal{B}_\mathrm{fin}\subseteq\mathcal{B}\subseteq\finsetsofthings$ are finite, this implies that $C\in\fin(K)$ because $\tilde{C}^*\subseteq\tilde{C}\subseteq C$.
\end{proof}

As an almost immediate consequence, we find that for finitary sets of desirable sets, the distinguishement between the different notions of coherence we consider, and their restrictions to $\finsetsofthings$, disappears. For coherence itself, this does require $\closure$ to be finitary though.

\begin{proposition}\label{prop:finitaryKbackandforth}
Consider a set of desirable sets $K\subseteq\setsofthings$ that is finitary. Then $K$ is finitely coherent if and only if it is finitely coherent in $\finsetsofthings$, and similarly for 2-coherence and 1-coherence. Furthermore, if $\closure$ is finitary, then $K$ is coherent if and only if it is coherent in $\finsetsofthings$.
\end{proposition}
\begin{proof}
As explained in the main text, in the paragraph following Definition~\ref{def:cohsodosQ}, coherence implies coherence in $\finsetsofthings$, and similarly for finite coherence, 2-coherence and 1-coherence. We therefore only need to prove the other directions. Since $K$ is finitary, we know from Proposition~\ref{prop:finitarywithfin} that $K=\fin(K)$. It therefore follows from Proposition~\ref{prop:fromKtofinK} that if $K$ is finitely coherent in $\finsetsofthings$, it is also finitely coherent, and similarly for 2-coherence and 1-coherence. Finally, assume that $K$ is coherent in $\finsetsofthings$ and $\closure$ is finitary. Then $K$ is also finitely coherent in $\finsetsofthings$ (because coherence in $\finsetsofthings$ implies finite coherence in $\finsetsofthings$). Since $K=\fin(K)$, it therefore follows from Proposition~\ref{prop:fromKtofinK} that $K$ is coherent.
\end{proof}

Taking into account our earlier equivalences between coherence in $\finsetsofthings$, finite coherence in $\finsetsofthings$ and 2-coherence in $\finsetsofthings$, this directly leads to the following similar results for the case $\mathcal{Q}=\setsofthings$, be it for finitary sets of desirable sets only.

\begin{corollary}\label{cor:fincohiffcoh}
If $\closure$ is finitary, a finitary set of desirable sets $K\subseteq\setsofthings$ is coherent if and only if it is finitely coherent.
\end{corollary}
\begin{proof}
We know from Proposition~\ref{prop:finitaryKbackandforth} that $K$ is coherent if and only if it is coherent in $\finsetsofthings$, and that $K$ is finitely coherent if and only if it is finitely coherent in $\finsetsofthings$. The result therefore follows from Corollary~\ref{cor:fincohiffcoh:finitecase}.
\end{proof}

\begin{corollary}\label{cor:fincohiffbincoh}
If $\closure$ is incremental, a finitary set of desirable sets $K\subseteq\setsofthings$ is finitely coherent if and only if it is 2-coherent.
\end{corollary}
\begin{proof}
We know from Proposition~\ref{prop:finitaryKbackandforth} that $K$ is finitely coherent if and only if it is finitely coherent in $\finsetsofthings$, and that $K$ is 2-coherent if and only if it is 2-coherent in $\finsetsofthings$. The result therefore follows from Proposition~\ref{fincohifbincoh:finitecase}.
\end{proof}

\begin{corollary}\label{cor:cohiffbincoh}
If $\closure$ is finitary and incremental, a finitary set of desirable sets $K\subseteq\setsofthings$ is coherent if and only if it is 2-coherent, and if and only if it is finitely coherent.
\end{corollary}
\begin{proof}
Immediate consequence of Corollaries~\ref{cor:fincohiffcoh} and~\ref{cor:fincohiffbincoh}.
\end{proof}

What is nice about these results is that if we combine them with our representation result in Theorem~\ref{theo:infinite:coherentrepresentation:twosided}, we see that for finitary sets of desirable sets, representation in terms of a set of coherent sets of desirable things can be obtained under weaker conditions than coherence. On the one hand, for finitary incremental closure operators, every set of desirable sets $K$ that is finitary and 2-coherent will be of the form $K_\setofdesirsets$, for some non-empty $\setofdesirsets\subseteq\desirsets$. On the other hand, for finitary closure operators, every set of desirable sets $K$ that is finitary and finitely coherent will be of the form $K_\setofdesirsets$. Unfortunately though, these results only provide sufficient conditions for such a representation, because even for finitary incremental closure operators, sets of desirable sets of the form $K_\setofdesirsets$ need not be finitary.

\begin{example}
We consider the case of desirable gambles, as discussed in Example~\ref{ex:gambles}, with $\closure=\posi$, $A_\mathrm{not}\coloneqq\{f\in\mathcal{G}(\mathcal{X})\colon f\leq0\}$ and $A_\mathrm{des}$ as in Equation~\eqref{eq:Adesgambles}. We focus on a simple version with two states only: $\states=\{a,b\}$. For any $n\in\naturals$, we let $f_n\in\mathcal{G}(\states)$ be the gamble defined by $f_n(a)\coloneqq-\nicefrac{1}{n}$ and $f_n(b)\coloneqq 1$, and use it to define a set of desirable gambles
\begin{equation*}
D_n\coloneqq\{f\in\mathcal{G}(\states)\colon f\geq\alpha f_n\text{ with }\alpha\in\reals_{\geq0}\}\setminus\{0\}.
\end{equation*}
It is a fairly simple exercise to show that each of these sets of desirable gambles is coherent. Now let $A\coloneqq\{f_n\colon n\in\naturals\}$. Then for all $n\in\naturals$, since $f_n\in A\cap D_n$, we know that $A\cap D_n\neq\emptyset$ and therefore, that $A\in K_{D_n}$. So we see that $A\in K_\setofdesirsets$, with $\setofdesirsets\coloneqq\{D_n\colon n\in\naturals\}$. However, there is no finite subset $B$ of $A$ such that $B\in K_\setofdesirsets$, because $f_m\notin D_n$ for all $n>m$.
\end{example}

The only simpler condition we have seen that is necessary and sufficient for coherence, and hence for a representation in the style of Theorem~\ref{theo:infinite:coherentrepresentation:twosided}, is 1-coherence for unitary closure operators, because that did not involve $K$ being finitary. Combining Theorem~\ref{theo:infinite:coherentrepresentation:twosided} with Proposition~\ref{prop:unitarycl} indeed immediately yields the following simplification of Theorem~\ref{theo:infinite:coherentrepresentation:twosided}.

\begin{theorem}\label{theo:infinite:1coherentrepresentation:twosided}
If $\closure$ is unitary, a set of desirable sets $\sodos$ is 1-coherent if and only if there is a non-empty set $\setofdesirsets\subseteq\desirsets$ of coherent sets of desirable things such that $\sodos=\sodos[\setofdesirsets]$. One such set $\setofdesirsets$ is $\mathcal{D}_K$, and the largest such set $\setofdesirsets$ is $\desirsets(\sodos)\coloneqq\cset{\desirset\in\desirsets}{\sodos\subseteq\sodos[\desirset]}$.
\end{theorem}
\begin{proof}
For unitary closure operators, we know from Proposition~\ref{prop:unitarycl} that coherence is equivalent to 1-coherence. The result therefore follows directly from Theorem~\ref{theo:infinite:coherentrepresentation:twosided}.
\end{proof}

There is however also another way in which we can obtain a representation in the style of Theorem~\ref{theo:infinite:coherentrepresentation:twosided} under conditions that are simpler than---yet still equivalent to---coherence, which consists in focussing on finite desirable sets only. That is, by considering sets of desirable sets in $\finsetsofthings$ that are coherent in $\finsetsofthings$. For those, we have already established several conditions that are simpler yet equivalent to coherence, so simplifying coherence is not the issue here. Instead, what's missing is a representation result for such models. In fact, it is not clear what such a representation would look like, since models of the form $K_\setofdesirsets$ are subsets of $\setsofthings$ but not of $\finsetsofthings$. This is easily fixed though because we can simply consider the restriction to $\finsetsofthings$. To that end, with any set of desirable things $D$, we associate the restriction
\begin{equation*}
K^\mathrm{fin}_D\coloneqq\{A\in\finsetsofthings\colon A\cap D\neq\emptyset\}=K_D\cap\finsetsofthings
\end{equation*}
of $K_D$ to $\finsetsofthings$, and with every non-empty set $\setofdesirsets$ of sets of desirable things, the restriction 
\begin{equation*}
K^\mathrm{fin}_\setofdesirsets\coloneqq\bigcap_{D\in\setofdesirsets}K^\mathrm{fin}_D=K_\setofdesirsets\cap\finsetsofthings
\end{equation*}
of $K_\setofdesirsets$ to $\finsetsofthings$. The question then is whether a result similar to Theorem~\ref{theo:infinite:coherentrepresentation:twosided} can be obtained here as well. That is, for any set of desirable sets in $\finsetsofthings$, is coherence in $\finsetsofthings$ equivalent with having a representation of this form, with $\setofdesirsets$ a set of coherent sets of desirable sets? Our next result shows that this is true for finitary closure operators; coherence in $\finsetsofthings$ can even be replaced by finite coherence in $\finsetsofthings$. The proof is fairly simple, and essentially consists in applying Theorem~\ref{theo:infinite:coherentrepresentation:twosided} to $\fin(K)$, which is coherent due to Proposition~\ref{prop:fromKtofinK}, and observing that $\fin(K)\cap\finsetsofthings=K$.

\begin{theorem}\label{theo:finfinrepresentation}
If $\closure$ is finitary, then a set of desirable sets $K\subseteq\finsetsofthings$ is finitely coherent in $\finsetsofthings$ if and only there is a non-empty set $\setofdesirsets\subseteq\desirsets$ of coherent sets of desirable things such that $\sodos=\sodos[\setofdesirsets]^\mathrm{fin}$. The same is true if we replace finite coherence in $\finsetsofthings$ by coherence in $\finsetsofthings$.
\end{theorem}
\begin{proof}
If $K$ is finitely coherent in $\finsetsofthings$, then since $\closure$ is finitary, it follows from Proposition~\ref{prop:fromKtofinK} that $\fin(K)$ is coherent and $\fin(K)\cap\finsetsofthings=K\cap\finsetsofthings=K$, where the last equality holds because $K\subseteq\finsetsofthings$. Since $\fin(K)$ is coherent, Theorem~\ref{theo:infinite:coherentrepresentation:twosided} implies that there is a non-empty set $\setofdesirsets\subseteq\desirsets$ such that $\fin(K)=K_\setofdesirsets$. Since $\fin(K)\cap\finsetsofthings=K$, this implies that $K=K_\setofdesirsets\cap\finsetsofthings=K_\setofdesirsets^\mathrm{fin}$. The same is of course true if $K$ is coherent in $\finsetsofthings$, because coherence in $\finsetsofthings$ implies finite coherence in $\finsetsofthings$.

Conversely, assume that $K=K_\setofdesirsets^\mathrm{fin}$ for some non-empty set $\setofdesirsets\subseteq\desirsets$. It then follows from Theorem~\ref{theo:infinite:coherentrepresentation:twosided} that $K_\setofdesirsets$ is coherent, which trivially implies that $K_\setofdesirsets\cap\finsetsofthings$ is coherent in $\finsetsofthings$, and therefore definitely finitely coherent in $\finsetsofthings$. Since $K=K_\setofdesirsets^\mathrm{fin}=K_\setofdesirsets\cap\finsetsofthings$, it follows that $K$ is finitely coherent in $\finsetsofthings$.
\end{proof}

Combining this result with Proposition~\ref{fincohifbincoh:finitecase}, we obtain a similar representation theorem for 2-coherence in $\finsetsofthings$, provided that $\closure$ is also incremental.

\begin{theorem}\label{theo:incrementalfinfinrepresentation}
If $\closure$ is finitary and incremental, then a set of desirable sets $K\subseteq\finsetsofthings$ is 2-coherent in $\finsetsofthings$ if and only there is a non-empty set $\setofdesirsets\subseteq\desirsets$ of coherent sets of desirable things such that $\sodos=\sodos[\setofdesirsets]^\mathrm{fin}$.
\end{theorem}
\begin{proof}
Immediate consequence of Theorem~\ref{theo:finfinrepresentation} and Proposition~\ref{fincohifbincoh:finitecase}.
\end{proof}

Combining Theorem~\ref{theo:finfinrepresentation} with Proposition~\ref{prop:unitarycl}, finally, yields a similar result for 1-coherence in $\finsetsofthings$, provided that $\closure$ is unitary.

\begin{theorem}\label{theo:unitaryfinfinrepresentation}
If $\closure$ is unitary, then a set of desirable sets $K\subseteq\finsetsofthings$ is 1-coherent in $\finsetsofthings$ if and only there is a non-empty set $\setofdesirsets\subseteq\desirsets$ of coherent sets of desirable things such that $\sodos=\sodos[\setofdesirsets]^\mathrm{fin}$.
\end{theorem}
\begin{proof}
For unitary closure operators, we know from Proposition~\ref{prop:unitarycl} that coherence in $\finsetsofthings$ is equivalent to 1-coherence in $\finsetsofthings$. Since every unitary closure operator is also finitary, the result therefore follows directly from Theorem~\ref{theo:finfinrepresentation}.
\end{proof}

Since many closure operators are both finitary and incremental, the most important of these three results is arguably Theorem~\ref{theo:incrementalfinfinrepresentation}; in particular, it allows us to recover and extend some earlier results for the case of vector spaces.

\begin{example}
We already explained in Example~\ref{ex:gambles} that representation results in the style of Theorem~\ref{theo:infinite:coherentrepresentation:twosided} have already been obtained for the case where $\things$ is a vector space---such as the set $\mathcal{G}(\mathcal{X})$ of all gambles on a state space $\mathcal{X}$---and $\closure=\posi$, but that these results took desirable sets to be finite, and replaced axiom~\ref{ax:sodos:cl} with a simpler version. We can now be more specific about this: the results we were referring to correspond to a special case of Theorem~\ref{theo:incrementalfinfinrepresentation}, with $\closure=\posi$---which is both finitary and incremental---and $\things$ either a vector space or, more specifically, a set of gambles $\mathcal{G}(\mathcal{X})$, and the simpler axiom we were referring to is then of course axiom~\ref{ax:sodos:clbinQfin}, which in this context, for $K\subseteq\finsetsofthings$, boils down to the following condition:
\begin{enumerate}[label=~~~%$\mathrm{K}_{5\mathrm{posi}}^\mathrm{fin}$.
,ref=$\mathrm{K}_{5\mathrm{posi}}^\mathrm{fin}$,leftmargin=*]
\item\label{ax:sodos:clposiQfin} if $A,B\in K$ and, for all $a\in A$ and $b\in B$, $\lambda_a,\lambda_b\in\reals_{\geq0}$ are such that $\lambda_a+\lambda_b>0$, then also $\{\lambda_a a+\lambda_b b\colon a\in A, b\in B\}\in K$.
\end{enumerate}
That said, Theorem~\ref{theo:incrementalfinfinrepresentation} is of course much more broadly applicable than the corresponding results in~\cite{debock2018,pmlr-v103-de-bock19b}: any set $\things$ will do, and it applies to any $\closure$ that is finitary and incremental. Since the convex hull operator $\chull$ is both finitary and incremental, this implies that Theorem~\ref{theo:incrementalfinfinrepresentation} can for instance be applied to the settings that we discussed in Examples~\ref{ex:nonlineargambles} and~\ref{ex:lotteriesandsuch}. In all of these settings, we can therefore replace~\ref{ax:sodos:cl} by~\ref{ax:sodos:clbinQfin} and still obtain a representation in the style of Theorem~\ref{theo:infinite:coherentrepresentation:twosided}, provided we focus on finite desirable sets only.
\end{example}

\begin{example}\label{ex:totalorders}
For our final example, we return to the setting of preferences, in Example~\ref{example:preferencesasthings}, and show that we can use it to represent sets of desirable sets of preferences in terms of strict total orders. To that end, let $\closure=\trans$ be the transitive closure operator of Example~\ref{example:preferencesasthings} and let $A_\mathrm{des}\coloneqq\emptyset$ and $A_\mathrm{not}\coloneqq\{(o,o)\colon o\in\options\}$. A set of desirable preferences $D$ is then coherent if and only if it is irreflexive and transitive, here expressed as $(o,o)\notin D$ and $(o_1,o_2),(o_2,o_3)\in D\Rightarrow(o_1,o_3)\in D$. Since $\trans$ is a finitary incremental closure operator, it therefore follows from Theorem~\ref{theo:incrementalfinfinrepresentation} that a set of desirable sets of preferences $K\subseteq\mathcal{P}_\mathrm{fin}(\mathcal{T}_\mathcal{O})$ is  2-coherent in $\mathcal{P}_\mathrm{fin}(\mathcal{T}_\mathcal{O})$ if and only if it is of the form $K^\mathrm{fin}_\setofdesirsets$, where each $D\in\setofdesirsets$ represents a binary relation on $\options$ that is irreflexive and transitive. The axiom~\ref{ax:sodos:clbinQfin} that is imposed by 2-coherence furthermore simplifies as follows:
\begin{enumerate}[label=~~~%$\mathrm{K}_{5\mathrm{posi}}^\mathrm{fin}$.
,ref=$\mathrm{K}_{5\mathrm{posi}}^\mathrm{fin}$,leftmargin=*]
\item\label{ax:sodos:clposiQfin} if $A,B\in K$ and, for all $a\in A$ and $b\in B$, $o_{ab}\in\trans(\{a,b\}$, then also $\{o_{ab}\colon a\in A, b\in B\}\in K$,
\end{enumerate}
where, for each $a=(a_1,a_2)\in A$ en $b=(b_1,b_2)\in B$, we have that
\begin{equation*}
\trans(\{a,b\})
\coloneqq
\begin{cases}
\{(a_1,a_2),(b_1,b_2)\} &\text{ if }a_2\neq b_1\text{ and }a_1\neq b_2\\
\{(a_1,a_2),(b_1,b_2),(a_1,b_2)\} &\text{ if }a_2=b_1\text{ and }a_1\neq b_2\\
\{(a_1,a_2),(b_1,b_2),(b_1,a_2)\} &\text{ if }a_2\neq b_1\text{ and }a_1= b_2\\
\{(a_1,a_2),(b_1,b_2),(a_1,b_2),(b_1,a_2)\} &\text{ if }a_2= b_1\text{ and }a_1= b_2\\
\end{cases}
\end{equation*}

For the representing binary relations to be strict total orders, we need them be connected as well, which here translates to the requirement that for any $o_1,o_2\in\options$ such that $o_1\neq o_2$, either $(o_1,o_2)\in D$ or $(o_2,o_1)\in D$. If we let
\begin{equation*}
\mathcal{A}_\mathrm{tot}\coloneqq\{\{(o_1,o_2),(o_2,o_1)\}\colon o_1,o_2\in\options,o_1\neq o_2\},
\end{equation*}
this corresponds to requiring that $A\cap 
D\neq\emptyset$ for all $A\in \mathcal{A}_\mathrm{tot}$ and $D\in\setofdesirsets$, or equivalently, that $\mathcal{A}_\mathrm{tot}\subseteq K_\setofdesirsets$. So we conclude that a set of desirable sets of preferences $K\subseteq\mathcal{P}_\mathrm{fin}(\mathcal{T}_\mathcal{O})$ can be represented by a set of total strict orders if and only if it is 2-coherent in $\mathcal{P}_\mathrm{fin}(\mathcal{T}_\mathcal{O})$ and includes $\mathcal{A}_\mathrm{tot}$. 
\end{example}

In summary, we see that focussing on finite desirable sets allows us to obtain a wide variety of necessary and sufficient conditions for representation. All of them require that $\closure$ is at least finitary though. If we do allow for infinite desirable sets, we know from Theorem~\ref{theo:infinite:coherentrepresentation:twosided} that coherence provides a necessary and sufficient condition for non-finitary closure operators as well, but this condition can then not be further simplified---at least not with our results---without giving up necessity.

\section{Where to go from here}\label{sec:conclusion}

The main contribution of this paper, I hope, has been to show that the theory of desirable gambles can be generalised from gambles to things, without giving up on the main ideas and results. That said, I have only done so up to some extent. That is, I have focussed on extending the notion of a coherent set of desirable gambles, and that of a coherent set of desirable sets of gambles, and on the connection between these two. There is more to the theory of desirable gambles though, and it remains to be seen to which extent these other aspects of the theory can be similarly extended as well. If the theory of desirable things is to be a proper generalisation of that of desirable gambles, as I hope it will, investigating these other aspects will be essential. To end this paper, I therefore give an overview of some of the aspects that remain to be explored, and in some cases, my initial thoughts on how these could be extended from gambles to things. I hope they can provide some inspiration for others wishing to further extend the framework here proposed.

\subsubsection*{Conservative inference}

One of the features of the theory of desirable gambles that makes it important for imprecise probabilities, is that it provides a natural way to do conservative inference: starting from an initial assessment (of either desirable gambles or desirable sets of gambles), if it is at all possible to extend it to a coherent model, then there will always be a unique smallest such coherent extension, called the natural extension~\cite{walley1991,debock2018}. That such a natural extension exists, is because coherence is preserved under taking intersections: the unique smallest coherent extension will therefore simply be the intersection of all possible coherent extensions. Although we haven't explored this feature explicitely, it follows from our results that the same ideas continue to hold for desirable things as well; indeed, as we have seen in Proposition~\ref{prop:intersectionofcoherentKsiscoherent}, coherence is still preserved under taking intersections. It should therefore be possible to develop and study a notion of natural extension here as well, similarly to what has been done for desirable gambles.

\subsubsection*{Additional axioms}

The closure operator that we employ in our definitions of coherence is both a strength and a weakness of the proposed approach. On the one hand, its generality allows us to apply this framework to a large variety of settings. On the other hand, however, having to put all of the specific inference aspects of the considered setting into this single operator, can lead to rather untransparent axiomatisations. This provides a host of possibilities for further exploration. 

On the one hand, it would be interesting to explore to what extent we can consider several closure operators at once, each of which imposes its own inference principle. We can try to do this for each type of model seperately, but we could also try to do this---and that is likely to be much more difficult---in such a way that the established connections between the two types of models continue to hold.

Another approach could be to stick with a single closure operator, but impose additional properties by means of other types of axioms. This was for example illustrated in Example~\ref{ex:totalorders}, where a simple extra condition guaranteed that the representing binary relations were strict total orders. More generally, I think the aim should be to come up with additional axioms that, when combined with coherence, allow for representation results where the representing models satisfy additional properties. Several successful examples of such an approach have already been obtained for desirable gambles~\cite{pmlr-v103-de-bock19b,ipmu2020debock:arxiv}, but it remains to be seen to which extent this is possible for desirable things as well.

\subsubsection*{Logic}

As hinted at in Footnote~\ref{footnote:logic}, the idea of representing inference principles with closure operators is used in (abstract) logic as well. Coherent sets of desirable things then essentially correspond to so-called closed theories. Results---or perhaps rather ideas---from abstract logic could therefore usefully be employed in the study of sets of desirable things.\footnote{Consider for example the well known result that coherent sets of desirable gambles can be represented in terms of maximal sets of desirable gambles. One might wonder to which extent this generalises to the setting of things. In abstract logic, this corresponds to the question of whether closed theories can be represented in terms of so-called maximal theories; a question that has been thoroughly studied in that field.} Sets of desirable sets of things, on the other hand, do not seem to have an analogue in abstract logic. Nevertheless, sets of desirable sets of things can be given a logical interpretation, which we've recently explored~\cite{cooman2023:things:logic:arxiv}.

\subsubsection*{Desirable preferences}

Traditionally, the link between desirable gambles and decision making relies heavily on the linear utility assumption, and the fact that gambles are elements of a vector space. That one gamble is preferred to another, can then be taken to be equivalent to the difference between these two gambles being desirable, at which point we can use desirability to model preferences. For general things, where such assumptions are not necessarily made, this is of course no longer possible. Nevertheless, as illustrated in Examples~\ref{example:preferencesasthings} and~\ref{ex:totalorders}, it is still possible to consider preferences---and hence decision making---without such assumptions, simply by directly considering the desirability of preferences. That is, to let the set of all things be the set of all preferences, calling a preference desirable if our subject has that preference, and to then impose inference principles directly on these preferences. Further exploration of this idea is definitely needed though to develop it into a full blown theory of decision making.

\subsubsection*{Choice functions}

One of the most general, and arguably also most intuitive ways in which the theory of desirable gambles has been connected with decision making, is through the connection between coherent sets of desirable sets of gambles on the one hand, and coherent choice functions on the other. In fact, these two types of models have been shown to be equivalent~\cite{pmlr-v103-de-bock19b}. The advantage of this connection is that we can combine the intuitive language of the latter---which involves statements about choices---with the mathematical power of the former, in the form of representation theorems such as the ones presented in this paper. In particular, this has lead to axiomatic characterisations for the use of various types of decision rules, including maximising expected utility, E-admissibility and maximality~\cite{pmlr-v103-de-bock19b,ipmu2020debock:arxiv,seidenfeld2010}. The results in this paper therefore beg the question whether a similar connection with choice functions is possible also in our more general setting. It think it is. In particular, I think it should be possible to establish a connection between coherent sets of desirable sets of preferences on the one hand, and coherent choice functions on the other, where the latter now choose between abstract options, rather than between gambles. In general, however, I don't think they will be equivalent. Instead, I think that choice functions will constitute a special case. Nevertheless, under suitable additional conditions, a complete equivalence could still be achievable.

\subsubsection*{Nonlinear operators}

One of the main reasons why coherent sets of desirable gambles have served as the mathematical underpining of many of the theoretical developments in the field of imprecise probabilities, is because, as explained in the introduction, many well known imprecise probability models correspond to special cases. Preference relations provide one example of such models, which we already discussed above, but an equally important class of such special cases are nonlinear operators, including coherent lower and upper expextations (or previsions) and various types of nonlinear set functions~\cite{walley1991,walley2000}.

Given the developments in this paper, it makes sense to wonder whether coherent sets of desirable things can similarly be connected with nonlinear operators. For general sets of things, this seems to make little sense though. To be able to even talk about (non)linearity, we will likely need to require that the things we consider should constitute a (subset of some) vector space. But once we do, it seems to me that it should be feasible to establish similar connections. In fact, even for the case of gambles, much remains to be done here, because our move to general closure operators opens up the possibility of establishing connections with other, more general types of nonlinear operators than the ones that are traditionally considered. Some examples of such connections have already been established in the work of Miranda and Zaffalon~\cite{miranda2022:nonlinear:desirability}, but other remain to be explored. If we adopt the convex hull as closure operator, for example, I expect that coherent sets of desirable gambles can be connected to convex lower and upper previsions and convex risk measures~\cite{Follmer2002,PELESSONI2005297}.
\subsubsection*{Multivariate models}

A final feature of desirable gambles that I would like to point out, is their use in a multivariate context: marginalisation, conditioning and independence, for example, are concepts that have been succesfully applied, not only to sets of desirable gambles~\cite{couso2011:desirable,cooman:2012:indnatexdesirs}, but also to sets of desirable sets of gambles~\cite{VANCAMP2023390}. This has for example made it possible to apply sets of desirable gambles in the context of credal networks~\cite{DEBOCK2015178}---imprecise generalisations of Bayesian networks. Staying with the case of gambles,  similar ideas can be developed for closure operators other than the standard $\posi$ operator as well. In fact, for  marginalisation and conditioning, this has already been explored~\cite{miranda2022:nonlinear:desirability}. The concept of independence remains to be extended to the setting of gambles with general closure operators though, as does the related notion of independent natural extension. The same is true for sets of desirable sets of gambles with general closure operators, for which even marginalisation and conditioning have yet to be studied.

Extending these ideas from gambles to arbitrary things seems impossible, but if the things in question have sufficient structure, this does seem feasible. For example, as one of the reviewers of an earlier conference version kindly suggested, we could consider uncertain things: mappings from a state space to a set of things; gambles and horse lotteries, for example, are specific instances of such uncertain things. These uncertain things are just special types of things, so the framework here presented can be applied, but they also provide sufficient structure to allow for a multivariate treatment.

%The page limit is 5 to 8 pages
\subsection*{Acknowledgements}
The main ideas in this paper were first presented during an invited talk at the EPIMP Inaugural Conference in October 2021, to which I was kindly invited by, and for which I received support from, Jason Konek. He gave me a free choice of topic, so I decided to come up with some new material for the occasion. One of the ideas I ended up presenting there, and with hindsight perhaps the most important one, was that of applying the notion of desirability to arbitrary things, using abstract closure operators to capture inference~\cite{debock2021:youtubeEPIMP}. I still vividly recall mentioning to some of the attendees that were interested in reading more about this idea, that it would surely take me at most a couple of weeks to properly write it down. Little did I know at the time that I would end up being so consumed with other things that it would take me sixteen months; I thank those people for their patience. I am also grateful to Arthur van Camp and Catrin Campbell-Moore, with whom I had several interesting discussions about choice functions and sets of desirable sets of gambles in the weeks leading up to the EPIMP conference. And when it comes to choice functions in general, my collaborations with Gert de Cooman on that topic~\cite{debock2018,pmlr-v103-de-bock19b,debock2000:festschriftTeddy} have served as a source of inspiration for the work here presented too. A (reduced) conference version of this paper has also benefitted from the constructive feedback of three anonymous reviewers, as well as that of Erik Quaeghebeur.  As for the financial side of things: work on this project was supported by my BOF Starting Grant “Rational decision making under uncertainty: a new paradigm based on choice functions”, number 01N04819.

\end{document}